 \documentclass[iicol,sn-basic]{sn-jnl}

\usepackage{graphicx}%
\usepackage{multirow}%
\usepackage{amsmath,amssymb,amsfonts}%
\usepackage{amsthm}%
\usepackage{mathrsfs}%
\usepackage[title]{appendix}%
\usepackage{xcolor}%
\usepackage{textcomp}%
\usepackage{manyfoot}%
\usepackage{booktabs}%
\usepackage{algorithm}%
\usepackage{algorithmicx}%
\usepackage{algpseudocode}%
\usepackage{listings}%

\usepackage{booktabs}
\usepackage{adjustbox}
\usepackage{colortbl}
\usepackage{graphicx}   
\usepackage{subcaption}
\usepackage{array}       
\usepackage{balance}

\theoremstyle{thmstyleone}%
\theoremstyle{thmstyletwo}%

\theoremstyle{thmstylethree}%
\begin{document}

\title[Contourlet Refinement Gate Framework for Thermal Spectrum Distribution Regularized Infrared Image Super-Resolution]{Contourlet Refinement Gate Framework for Thermal Spectrum Distribution Regularized Infrared Image Super-Resolution}


\author[1,5]{\fnm{Yang} \sur{Zou}}\email{archerv2@mail.nwpu.edu.cn}

\author[2]{\fnm{Zhixin} \sur{Chen}}\email{chenzhixin@akane.waseda.jp}

\author[1,5]{\fnm{Zhipeng} \sur{Zhang}}\email{zhipeng.zhang@mail.nwpu.edu.cn}

\author[3]{\fnm{Xingyuan} \sur{Li}}\email{xingyuan\_lxy@163.com}

\author[3]{\fnm{Long} \sur{Ma}}\email{malone94319@gmail.com}

\author*[4]{\fnm{Jinyuan} \sur{Liu}}\email{atlantis918@hotmail.com}

\author[1,5]{\fnm{Peng} \sur{Wang}}\email{peng.wang@nwpu.edu.cn}

\author[1,5]{\fnm{Yanning} \sur{Zhang}}\email{ynzhang@nwpu.edu.cn}

\affil[1]{\orgdiv{School of Computer Science}, \orgname{Northwestern Polytechnical University}, \orgaddress{\city{Xi'an}, \postcode{710129}, \country{China}}}

\affil[2]{\orgdiv{Graduate School of Information, Production and Systems}, \orgname{Waseda University}, \orgaddress{\postcode{8080135}, \state{Fukuoka}, \country{Japan}}}

\affil[3]{\orgdiv{School of Software}, \orgname{Dalian University of Technology}, \orgaddress{\city{Dalian}, \postcode{116024}, \country{China}}}

\affil[4]{\orgdiv{ School of Mechanical Engineering}, \orgname{Dalian University of Technology}, \orgaddress{\city{Dalian}, \postcode{116024}, \country{China}}}

\affil[5]{\orgname{The National Engineering Laboratory for Integrated Aero-Space-Ground-Ocean Big Data Application Technology}, \country{China}}

\abstract{Image super-resolution (SR) is a classical yet still active low-level vision problem that aims to reconstruct high-resolution (HR) images from their low-resolution (LR) counterparts, serving as a key technique for image enhancement. Current approaches to address SR tasks, such as transformer-based and diffusion-based methods, are either dedicated to extracting RGB image features or assuming similar degradation patterns, neglecting the inherent modal disparities between infrared and visible images. When directly applied to infrared image SR tasks, these methods inevitably distort the infrared spectral distribution, compromising the machine perception in downstream tasks. In this work, we emphasize the infrared spectral distribution fidelity and propose a Contourlet refinement gate framework to restore infrared modal-specific features while preserving spectral distribution fidelity. Our approach captures high-pass subbands from multi-scale and multi-directional infrared spectral decomposition to recover infrared-degraded information through a gate architecture. The proposed Spectral Fidelity Loss regularizes the spectral frequency distribution during reconstruction, which ensures the preservation of both high- and low-frequency components and maintains the fidelity of infrared-specific features. We propose a two-stage prompt-learning optimization to guide the model in learning infrared HR characteristics from LR degradation. Extensive experiments demonstrate that our approach outperforms existing image SR models in both visual and perceptual tasks while notably enhancing machine perception in downstream tasks. Our code is available at \href{https://github.com/hey-it-s-me/CoRPLE}{\texttt{https://github.com/hey-it-s-me/CoRPLE}}.}


\maketitle

\section{Introduction}\label{sec1}
Image super-resolution (SR) ~\cite{glasner2009super,dong2015image,lim2017enhanced} aims to enhance a LR image to its HR counterpart, bypassing the need for hardware modifications. It is pivotal in a wide range of applications, such as remote sensing image processing ~\cite{lei2017super,yang2015remote}, security surveillance ~\cite{shamsolmoali2019deep}, and medical imaging ~\cite{li2022transformer,zhang2021mr}. Studying infrared image processing in image SR is fundamental for tasks like segmentation~\cite{liu2023multi}, infrared object detection~\cite{liu2022target}, and pedestrian detection~\cite{alfred2023fully}. As a vital component of image SR, infrared imaging detects thermal radiation from objects and offers consistent performance under challenging conditions, functioning effectively throughout the day and night. The robustness of infrared images makes them indispensable for various applications where visibility might be compromised. 

Unfortunately, infrared sensors are inherently challenged by several limitations, including sensitivity to temperature variations, high noise levels, diminished spatial resolution, and constrained dynamic range ~\cite{ma2019infrared}. These inherent issues significantly degrade the quality of infrared images. Additionally, the diversity of images, the presence of occlusions, and interference from backgrounds further complicate infrared image processing. The longer wavelengths of infrared radiation, compared to visible light, contribute to lower spatial resolution and reduced detail in captured images ~\cite{ma2019infrared}. These factors collectively hinder essential tasks such as object detection~\cite{alfred2023fully,liu2022target}, tracking~\cite{yilmaz2006object, wu2013online, kristan2015visual}, and segmentation ~\cite{fu2019dual, pu2018graphnet}. Consequently, improving the resolution of infrared images to enhance contrast and detail is crucial for increasing the effectiveness and precision of these tasks.

\begin{figure}[!htp]
    \centering
    \includegraphics[width=1.0\linewidth]{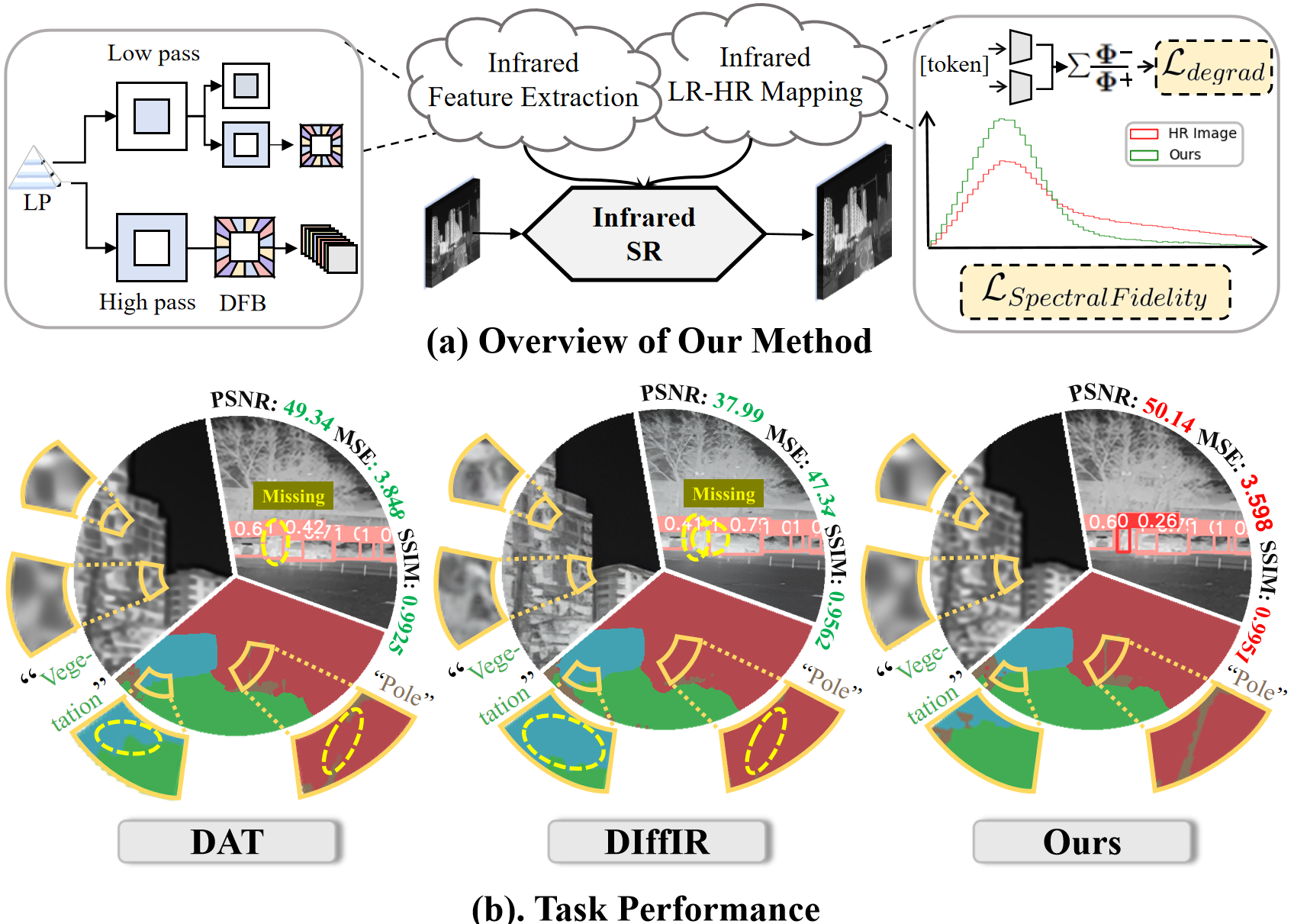}
    \caption{Algorithmic core design and performance evaluation of our method. In (a), our approach efficiently reconstructs high-quality infrared images by leveraging the multi-scale and multi-directional Contourlet Transform to extract quality-sensitive, infrared-specific high-pass subbands. We further learn the infrared LR-HR mapping using degradation and spectral fidelity loss, improving the fusion results and machine perception. In (b), we present a statistical objective comparison of our proposed method across super-resolution, object detection, and semantic segmentation tasks, benchmarking it against two SOTA methods and demonstrating its superior performance in the infrared domain.}
    \label{pics:shoutu}
\end{figure}

In recent years, numerous approaches have been developed to tackle image SR, from convolutional neural network (CNN)-based methods to diffusion-based innovations, driving significant advancements in the field. CNN-based methods ~\cite{dong2015image, dong2016accelerating,lim2017enhanced} often focus on complex architectural designs, utilizing spatially invariant kernels to capture local features. However, these methods fall short in effectively modeling pixel relationships and establishing long-range dependencies ~\cite{liang2021swinir}. To address these limitations, transformer-based methods ~\cite{mei2021image, zamir2022restormer,liang2021swinir} have been introduced, incorporating self-attention mechanisms to capture global context interactions. More recently, diffusion-based approaches~\cite{gao2023implicit,wu2024one,xiao2024single} have emerged, leveraging powerful pre-trained text-to-image (T2I) models such as Stable Diffusion (SD). These approaches fine-tune T2I models with adapters, allowing them to reconstruct HR images using LR images as the control signal.

In the end, modern image SR pipelines boil down to solving a set of core problems: extracting features, learning LR-HR image mapping, and reconstructing HR images. Considering most of the existing advances in this field are designed with visible light images, i.e., extracting visible light image features and assuming visible light image degradation patterns to learn LR-HR image mapping, we argue current image SR methods do not adequately address the unique characteristics of infrared images in this manner. In this regard, the outcomes of our previous work~\cite{licorple} are quite telling: infrared images have their unique degradation patterns, i.e., the quality of infrared images is more sensitive to high-frequency information due to their longer wavelengths and less susceptibility to atmospheric scattering, containing fewer high spatial frequency components. More seriously, the process of forward propagation in neural networks often results in the diminution of high-frequency details ~\cite{ma2019fusiongan, zhao2021efficient}. Therefore, in our previous work, inspired by the Contourlet Transform~\cite{do2005contourlet}, we prove that restoring the high-frequency subbands from infrared images notably improves the infrared image SR results. On top of that, we found regularizing the spectral frequency distribution during reconstruction is also critical for learning infrared LR-HR image mapping.

In this paper, we present an infrared spectral distribution-regularized SR framework dedicated to infrared image degradation and infrared-specific feature restoration from multi-scale and multi-directional infrared spectral decomposition. We are the first to emphasize the importance of preserving infrared spectral distribution fidelity for better learning of infrared LR-HR image mapping. To this end, we propose the Spectral Fidelity Loss, which constrains the distribution of high- and low-frequency thermal spectrum to maintain the fidelity of infrared-specific features. To better extract and restore quality-sensitive high-frequency features in infrared images, we leverage the multi-directional and multi-scale analysis capabilities of the Contourlet Transform, along with a global-local interactive attention block that efficiently captures the cross-window interactions of local patches to enhance the deep feature extraction of infrared images. Also, our proposed method is capable of understanding deep semantic information through a two-stage prompt learning strategy, which distinguishes right from wrong by guiding the optimization process with paired positive and negative textual prompts. 

While parts of the results in this paper were originally presented in the conference version~\cite{licorple}, this extended version expands upon our earlier work in several key aspects:

\begin{itemize}
\item  To better extract and restore quality-sensitive infrared high-frequency features, we introduce a prompt-learning Contourlet refinement gate framework tailored for infrared image SR to restore and enhance the high-frequency details from the multi-scale infrared spectra decomposition, crucial for reconstructing high-pass subbands-lacked infrared images.
\item  To better learn the infrared LR-HR image mapping, we propose the Spectral Fidelity Loss to constrain the distribution of high- and low-frequency thermal spectrum to maintain the fidelity of infrared-specific features, not only improving the fusion results but also notably enhancing the machine perception in downstream vision tasks.
\item  We conduct comprehensive experiments to verify that our method surpasses existing super-resolution algorithms and achieves state-of-the-art performance in various downstream tasks,i.e., detection and segmentation, setting a new paradigm in the area of infrared image super-resolution.
\end{itemize}

\section{Related Work}\label{sec2}
\subsection{Image Super-Resolution}
\textbf{GAN-based SR Method.}
Starting from SRCNN ~\cite{dong2014learning}, numerous deep learning-based Super-Resolution (SR) have been proposed. By employing more complex and realistic degradation models, super-resolution technologies like Real-ISR~\cite{liang2022efficient} can better adapt to and handle real-world image degradation issues, offering image reconstruction. Specifically, BSRGAN~\cite{zhang2021designing} and Real-ESRGAN~\cite{wang2022realesrgan} use a wide variety of training samples with different types of degradation, leveraging GANs to enhance the quality and resolution of low-resolution images. BSRGAN employs a randomly shuffling degradation modeling process, while Real-ESRGAN proposes a high-order degradation strategy. 

Nonetheless, GANs can suffer from training instability and may introduce artifacts, especially in extremely low-resolution inputs, which are evident in Real-ISR outputs. Subsequent methodologies like LDL~\cite{liang2022details} and DeSRA~\cite{xie2023desra} have mitigated these artifacts but struggle to reproduce natural details. Moreover, many methods face challenges in degradation modeling, which has a significant impact on the quality and effectiveness of SR. 

\textbf{Transformer-based SR methods.}
The Transformer model is originally designed for natural language processing. It has found applications in the field of computer vision tasks in recent years, such as image classification~\cite{dosovitskiy2020image,li2020lapar,liu2021transformer,liu2021swin,vaswani2021scaling,wu2020visual}, object detection~\cite{carion2020end,liu2020deep,liu2021swin,touvron2021training} and segmentation~\cite{cao2022swin,zheng2021rethinking,liu2021swin,wu2020visual}. Due to its exceptional performance, Transformer models have also been introduced for low-level vision problems such as image restoration~\cite{hong2024light, ma2023bilevel, zhang2021beyond}. 

Although IPT~\cite{chen2021pre} demonstrates the architecture's adaptability to various image processing tasks, it requires large datasets to reach its potential. VSR-Transformer~\cite{cao2021video} enhances the resolution of video frames through temporal and spatial features, but image features are still extracted from CNN. 
Liang et al. proposed SwinIR~\cite{liang2021swinir}, which harnesses the powerful feature extraction capabilities of the Swin Transformer, combining the strengths of both CNNs and Transformers to restore high-quality images in SR tasks. However, its network structure is largely borrowed from the Swin Transformer, which is designed for high-level vision tasks and is redundant for the SR problem. It computes self-attention (SA) within small fixed-size windows, preventing the utilization of long-range feature dependencies. 

In contrast, ELAN~\cite{zhang2022efficient} can compute SA in larger windows, and Restormer~\cite{zamir2022restormer} can learn long-range dependencies while maintaining computational efficiency. Recently, DAT~\cite{chen2023dual} proposed a new Transformer model for image SR that aggregates spatial and channel features in inter-block and intra-block dual manner. CAMixerSR~\cite{wang2024camixersr} integrates model accelerating and token mixer designing strategies by routing self-attention of varied complexities according to content complexities.

\textbf{Diffusion-based SR methods.}
The diffusion model (DM) has gained significant attention for its ability to generate high-quality, complex data distributions. In image-related tasks, DM has been particularly successful in image editing~\cite{avrahami2022blended}, inpainting~\cite{lugmayr2022repaint}, and deblurring~\cite{whang2022deblurring}, they can use a reduced number of sampling steps to expedite the generation process. In the area of SR, DM has been explored as well. SR3~\cite{saharia2022image} conditions a diffusion model on low-resolution (LR) images, gradually refining them to high-resolution (HR) images. SRDiff~\cite{li2022srdiff}  employs a residual prediction strategy to accelerate training speed and optimize the network more effectively, it also utilizes encoded LR information for noise prediction. 

More recently, IDM~\cite{gao2023implicit} introduced the implicit image function in the decoding part of the diffusion model, while OSEDiff~\cite{wu2024one} directly took the given low-quality image as the starting point for diffusion, eliminating the uncertainty associated with random noise. However, these methods still rely heavily on LR images and often overlook the use of other modalities, such as text, to provide additional information or priors for the enhancement process. 

\subsection{Text Prompt Image Processing}
The emergence of the Vision-Language Models enables us to manipulate or generate new images based on text descriptions. Vision-Language Models~\cite{radford2021learning, jia2021scaling, yao2021filip} generally adopt multiple objectives in downstream tasks to explore their synergy, leading to more robust models and improved performance in generating new images. For instance, DALL-E-2~\cite{ramesh2022hierarchical}, Imagen~\cite{saharia2022photorealistic} and Stable Diffusion~\cite{rombach2022high} all utilize diffusion models for text-to-image generation. DALL-E-2 significantly enhances sample quality by modulating the diffusion process with CLIP~\cite{radford2021learning} image embeddings instead of original text embeddings. Imagen further demonstrates the effectiveness of large pre-trained language models as text encoders for text-to-image generations. 

Additionally, the Stable Diffusion is publicly available and sufficiently compact. Various methodologies have been employed for image manipulation. StyleCLIP~\cite{patashnik2021styleclip} combines the generative capabilities of StyleGAN~\cite{karras2019style} with the vision-language abilities of CLIP~\cite{radford2021learning}, and DiffusionCLIP~\cite{kim2022diffusionclip} uses diffusion models alongside CLIP for image generation. Meanwhile, Prompt-to-Prompt~\cite{hertz2022prompt} and InstructPix2Pix~\cite{brooks2023instructpix2pix} are pre-trained and fine-tuning-free approaches. Prompt-to-Prompt edits images within a pre-trained diffusion model by modifying prompts, whereas InstructPix2Pix edits images based on input instructions. However, the use of text prompts in image SR has previously seen limited exploration. In our work, we investigate the application of text prompts in the context of infrared image SR.

\begin{figure*}[!htp]
    \centering
    \includegraphics[width=1.0\linewidth]{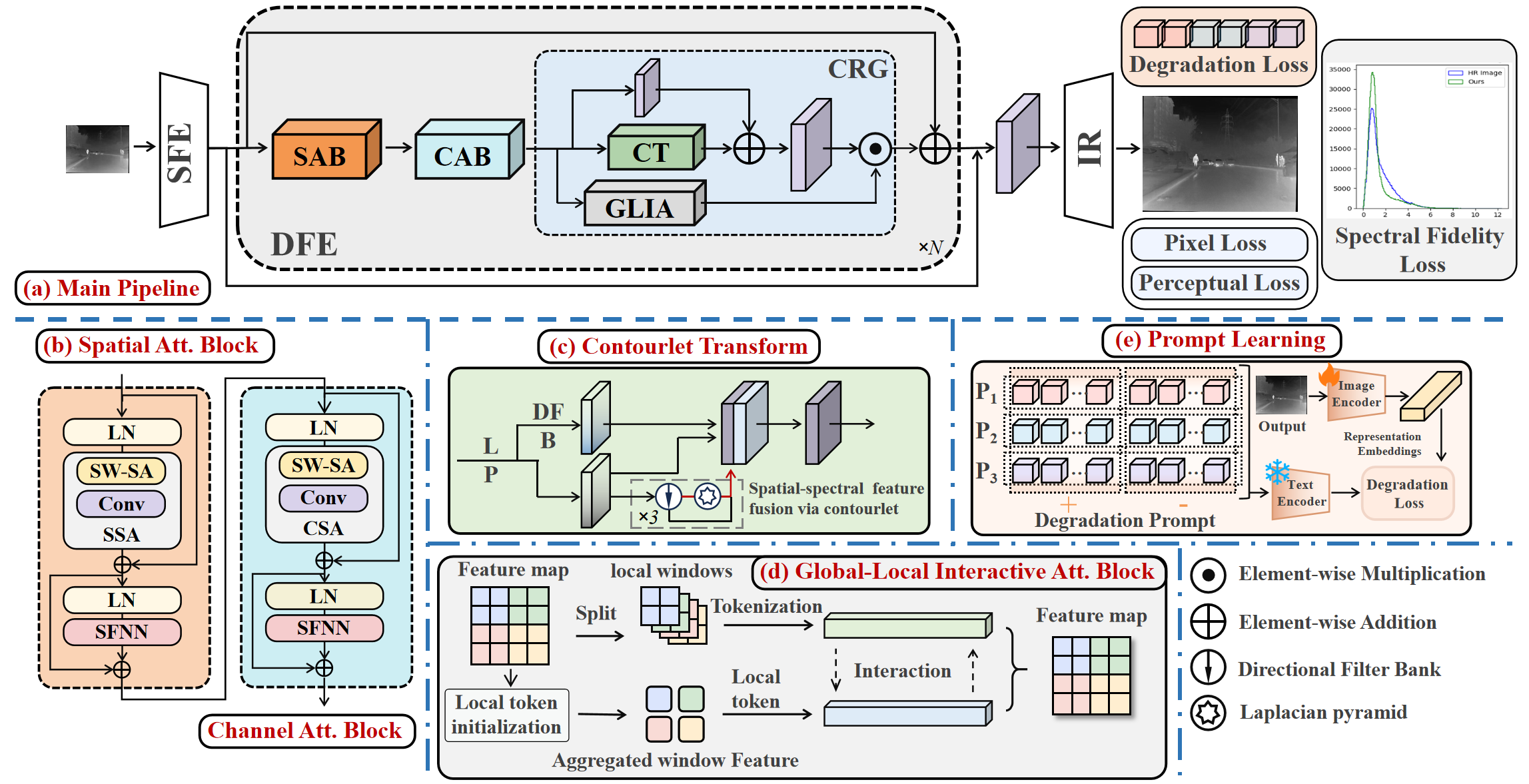}
    \caption{Overall architecture of our network. (a) The main pipeline consists of shallow feature extraction, deep feature extraction, and image reconstruction. The deep feature extraction contains the Spatial Attention Block (SAB), Channel Attention Block (CAB), and the Contourlet Refinement Gate (CRG). (b) The spatial Attention Block (SAB) and the Channel Attention Block (CAB). (c) The Contourlet Transform (CT). (d) The Global-Local Interactive Attention Block. (e) Prompt learning optimization.}
    \label{pics:cover}
\end{figure*}

\subsection{Infrared Image Enhancement}
With the successful development of deep learning,  especially the neural networks that have been recognized for their remarkable capabilities in SR tasks, many efforts have started investigating infrared image enhancement~\cite{marivani2020joint,yang2020deep, liu2024coconet}. Infrared image enhancement aims to improve the quality of images captured by infrared sensors. 

However, reconstructing edge details in infrared images presents a significant challenge~\cite{ma2021image, ma2019locality}. The defects of infrared imaging devices and external environmental factors often result in images with low contrast, unclear target edges, and poor visual effects ~\cite{liu2022target}. These challenges are exacerbated by the longer wavelengths of infrared radiation than visible light, leading to images with reduced spatial resolution and diminished detail~\cite{liu2023multi}. Marivani et al.~\cite{marivani2020joint} initially integrate sparse edge information from visible light images and combine it with interpretable sparse priors. Jun et al.~\cite{chen2024mdbfusion} propose a joint motion deblurring and fusion network for visible and infrared images, which effectively preserves texture details and enhances deblurring performance through the fusion of infrared images. 

Other researchers have developed modules capable of extracting high-frequency information from visible light images and using attention mechanisms to migrate this pattern information into the infrared feature domain~\cite{huang2021infrared,yang2020deep,patel2021thermisrnet,zou2020infrared}. 
Recently, our previous work~\cite{licorple} introduced a specialized Contourlet residual framework that captures the high-pass subbands of infrared images using the Contourlet Transform, notably improving the fusion performance.

\section{The Proposed Method}\label{sec3}
In this section, we detail our methodology by first elaborating on the motivation of the proposed method and then introducing the detailed network architecture. Afterward, the proposed spectral fidelity loss and degradation loss are presented.

\subsection{Motivation}
We consider the goal of image SR to involve extracting features, learning LR-HR image mappings, and subsequently reconstructing HR images ~\cite{wu2024one}. However, for the infrared image SR task, explicit supervision signals are lacking as guidance to learn infrared characteristics.

Current SR methods are predominantly developed and optimized for visible light images. As a result, these methods generally focus on extracting features from visible light images and training models to capture patterns such as textures, edges, and color gradients that are characteristic of RGB images. This feature extraction process is highly specialized to the unique properties of visible light, which differ significantly from those of infrared images. Additionally, these methods often assume similar degradation patterns between LR and HR images in the visible light domain, relying on predefined models of blurring, down-sampling, and noise that are specific to RGB data. Implicitly, these methods expect that the degradation and restoration processes for infrared images will follow the same patterns as in visible light, which is not the case due to the aforementioned distinct physical properties of infrared radiation.

Consequently, the reconstructed infrared images often suffer from artifacts and perform sub-optimally on downstream tasks such as detection and segmentation. We argue that current SR methods do not adequately address the unique characteristics of infrared images, as they fail to account for the different spectral and degradation behaviors inherent in infrared data.

To address these challenges, we propose an infrared spectral distribution-regularized SR framework. By introducing Spectral Fidelity Loss, which constrains the distribution of high- and low-frequency thermal radiation. Along with an infrared-specific prompt learning mechanism, our approach ensures the preservation of infrared-specific features, improving the learning of LR-HR mappings for infrared images. Furthermore, we incorporate a Contourlet refinement gate framework to restore and enhance high-frequency information specific to infrared images from multi-scale infrared spectral decomposition, critical for reconstructing infrared images that typically lack high-pass subbands.

\subsection{Architecture}
As illustrated in Figure~\ref{pics:cover} (a), our proposed network consists of three primary modules: the shallow feature extraction module, the deep feature extraction module, and the HR image reconstruction module. The process starts with the shallow feature extraction phase, where the LR input image is passed through a \(3 \times 3\) convolutional layer to preliminarily parse the image's basic features. Progressing deeper, the model moves into the deep feature extraction phase, which integrates spatial and infrared spectral features by leveraging spatial and channel attention blocks, along with the Contourlet refinement gate restoring the infrared-specific information. 

Finally, the HR image reconstruction module is responsible for reconstructing the HR image. In this phase, the combined features are upscaled using the pixel shuffle method ~\cite{shi2016real}, with convolutional layers aggregating the features into the final HR output.

\subsection{Spatial Attention Block}
As shown in Figure~\ref{pics:cover} (b), the Spatial Attention Block (SAB) operates by segmenting feature spaces into discrete spatial windows, applying self-attention within each to capture intricate spatial relationships. For an input feature matrix $X\in\mathbb{R}^{H \times W \times C}$, we employ learnable weights to generate query (Q), key (K), and value (V) matrices. These matrices are then subdivided into non-overlapping windows, with each window processed independently to emphasize localized feature interactions. The process involves dividing these matrices into multiple heads, enabling parallel processing of diverse feature aspects. 

The attention scores are computed through the dot product of the queries and keys, normalized with the softmax function to obtain attention weights for each position. The attention outputs from all heads are then concatenated and subjected to a linear projection, ensuring a comprehensive integration of spatially attentive features into a unified representation. 
 
\subsection{Channel Attention Block}
Different from the SAB, the Channel Attention Block (CAB) applies self-attention across channels for each window, capturing global channel relationships, as shown in Figure~\ref{pics:cover} (b). Given input $X\in\mathbb{R}$, we linearly project it to form query (Q), key (K), and value (V) matrices. These are then reshaped to $Q_c$, $K_c$, and $V_c$ with dimensions $\mathbb{R}^{HW \times C}$, and divided into multiple heads (h). Attention is computed for each head, using a learnable scaling factor ($\alpha$) for normalization:
\begin{equation}
Y^i_c = \text{softmax}\left(\frac{(Q^i_c)^T K^i_c}{\alpha}\right) \cdot V^i_c,
\end{equation}
\noindent yielding the final channel-wise attention output by concatenating and reshaping head outputs, akin to the spatial attention process.

\noindent \textbf{Spatial Feed-forward Neural Network.}
The Spatial Feed-Forward Neural Network (SFNN) improves spatial information processing and reduces channel redundancy through the integration of a spatial gating mechanism. By utilizing depth-wise convolution and element-wise multiplication, SFNN enhances efficiency within the architecture. The network splits the input feature map ($\hat{X}$) along the channel axis, processes each segment independently with convolutional and multiplicative operations, and subsequently merges the processed parts. This approach allows for more effective spatial information handling while maintaining computational efficiency:

\begin{equation}
\begin{split}
    &\hat{X}^{\prime} = \sigma(W^1_p \hat{X}), \quad [\hat{X}^{\prime}_1, \hat{X}^{\prime}_2] = \hat{X}^{\prime}, \\
    &\text{SFNN}(\hat{X}) = W^2_p (\hat{X}^{\prime}_1 \odot (W_d \hat{X}^{\prime}_2)),
\end{split}
\end{equation}
\noindent where $\odot$ denotes the element-wise multiplication, $W^1_p$ and $W^2_p$ are the weight matrix used for linear projection. The notation $\sigma$ represents the GELU function, and $W_d$ represents the depth-wise convolutional parameters. This ensures our model better preserves the spatial feature, which is important for the latter infrared-specific high-frequency feature decomposition.

\subsection {Contourlet Refinement Gate}
The Contourlet Refinement Gate (CRG) module shown in Figure~\ref{pics:cover} (a) blue part is designed to effectively capture quality-sensitive high-frequency details and leverage global-local feature interaction information by combining multi-scale, multi-directional contourlet decomposition with adaptive global-local feature interactions. The gating mechanism selectively emphasizes important features while suppressing redundancies, and the residual connection ensures the preservation of high-frequency information. 

\begin{figure}[!pt]
    \centering
    \includegraphics[width=0.7\linewidth]{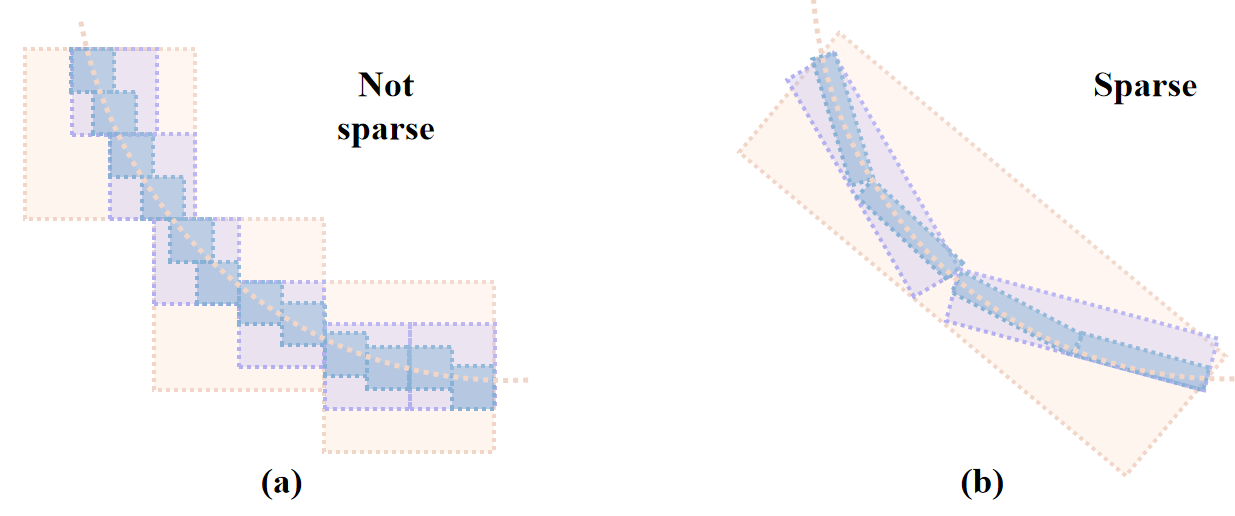}
    \caption{Visual demonstration of the sparsity for (a). traditional Wavelet Transform and (b). Contourlet Transform.}
    \label{pics:contourlet sparse}
\end{figure}

\noindent \textbf{Why Contourlet Transform?}
Our choice of the Contourlet Transform over the Wavelet Transform is driven by its superior capability to capture the multidimensional singularities commonly found in infrared images, such as edges, lines, and contours. While the Wavelet Transform offers satisfactory time-frequency localization, its square support makes it less effective at capturing high-dimensional features, and it struggles to provide a sparse representation~\cite{liu2020c}. 

Furthermore, the convolutional nature of the Wavelet Transform tends to be computationally expensive and lacks translation invariance, leading to issues like the Gibbs phenomenon~\cite{do2005contourlet}. In contrast, the Contourlet Transform extends the advantages of wavelets into higher dimensions, offering a more efficient sparse representation where smooth image contours are captured with fewer coefficients. This results in enhanced robustness across different scales and orientations, as illustrated in Figure~\ref{pics:contourlet sparse}.

\begin{figure}[!pt]
    \centering
    \includegraphics[width=0.65\linewidth]{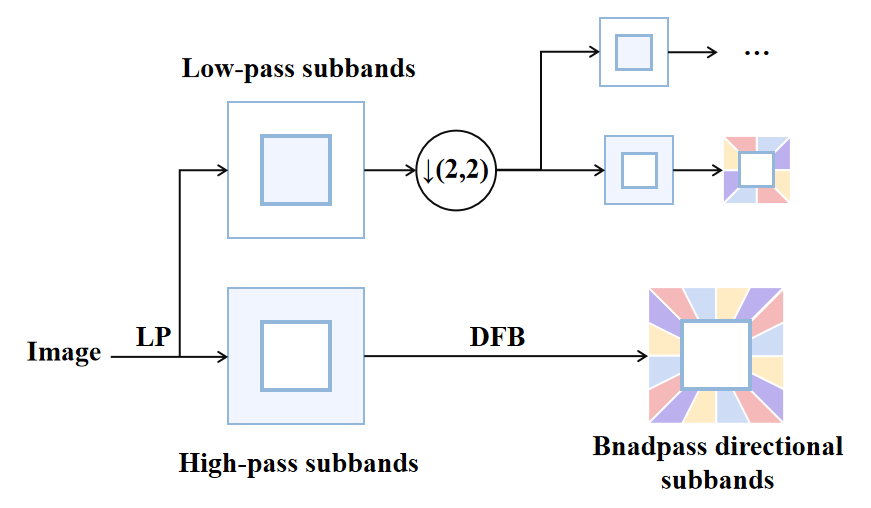}
    \caption{Architecture of the Contourlet Transform, The input feature is first decomposed by an LP filter to low- and high-pass subbands. Then, the high-pass subbands are decomposed into \(2^i\) directional subspaces through the DFB.}
    \label{pics:contourlet process}
\end{figure}

\noindent \textbf{Contourlet Transform.}
Upon extracting global deep features through the SAB and CAB blocks, we employ a Contourlet Transform block, as illustrated in Figure~\ref{pics:cover} (c) and Figure~\ref{pics:contourlet process}. This block starts with a Laplacian Pyramid decomposition of the deep features, denoted as \(\mathbf{X}\), segregating them into low and high-pass components. The low-frequency subband \(\mathbf{X_{\text{low}}}\), encapsulating the core structure and broad contours of the image, is procured through Gaussian filter down-sampling expressed as \(G_i(x, y) = (X_{i-1} * h)(2x, 2y)\), where \( G_i \) represents the image at the \( i^{th} \) level of the Gaussian pyramid, \( X_{i-1} \) is the feature from the previous level, \( h \) denotes the Gaussian filter, \( * \) is the convolution operation, and \( (2x, 2y) \) signifies the down-sampling process, capturing the general features across scales. 

The Laplacian layer \( L_i \), embodying the high-frequency subband \(\mathbf{X_{\text{high}}}\), is derived by:

\begin{equation}
    L_i(x, y) = X_{i-1}(x, y) - (G_i * h^T)(x, y),
\end{equation}

\noindent with \(L_i\) forming the \(i^{th}\) level of the Laplacian pyramid and \( h^T \) being the transposed Gaussian filter for reconstructing the preceding layer's features. The high-frequency subband \(\mathbf{X_{\text{high}}}\), ensuing from the LP decomposition, undergoes a subsequent refinement via the Directional Filter Bank decomposition:

\begin{equation}
    B_{l,k}(x, y) = (X_l * f_k)(x, y),
\end{equation}

\noindent where \( B_{l,k} \) denotes the subband for the \( l^{th} \) level and \( k^{th} \) direction, and \( f_k \) is the directional filter. The DFB decomposition dissects the image into directionally sensitive subbands, allowing the model to discern textural and edge details within the high-frequency domains of the infrared spectrum. This decomposition is iteratively applied to the low-pass subband at each LP level, recursively utilizing both LP and DFB to extract and refine high-frequency features across the infrared spectrum from coarse to fine. 

The resultant coefficients \(\mathbf{X}_{\text{spectral}}\) are a confluence of the multi-scale, multi-directional features, expressed as \(\mathbf{X}_{\text{spectral}} = \{ L_i(x, y) \} \cup \{ B_{l,k}(x, y) \}\), where \{\( L_i(x, y) \)\} is the set of all Laplacian pyramid layers, and \{\({ B_{l,k}(x, y) }\)\} is the set of all DFB decomposed subbands. The coefficients obtained from the Contourlet Transform capture both the essential low-frequency components and the sparse high-frequency details, providing a comprehensive representation of infrared image features for image reconstruction. 

After enhancing the spectral features using the Contourlet Transform, they are fused with the spatial features obtained from the CAB block through a residual connection, thereby enriching the high-frequency detail representation in the reconstructed infrared images.

\noindent \textbf{Global-Local Interactive Attention Block.}
Simultaneously, the deep features extracted from CAB are passed through a Global-Local Interactive Attention (GLIA) block illustrated in Figure~\ref{pics:cover} (d). The GLIA block enhances both local and global contextual information by employing hierarchical attention mechanisms that capture dependencies across different regions of the image. Let the input feature map be $\mathbf{X} \in \mathbb{R}^{H \times W \times d}$ with H, W, and d denote the height, width, and dimension, respectively. 

We start with local windows introduced in
Swin Transformer~\cite{liu2021swin} to partition the input feature map into \(n \times n\) local windows with \(n = \frac{H \times W}{k^2}\), where k is the window size. To efficiently model long-range dependencies, we use local tokens ($\mathbf{C}$) to summarize the information of the local windows. The local tokens are initialized by applying a \(3 \times 3\) convolution operation followed by a pooling operation, which reduces spatial resolution as \(\mathbf{C} = \text{Pooling}_{\text{H} \times \text{W} \to \text{n} \times \text{L}}(\text{Conv}(\mathbf{X}))\), where \(\text{L}\) is the number of local tokens per window. 

\begin{figure*}[!hpt]
    \centering
    \includegraphics[width=1.0\linewidth]{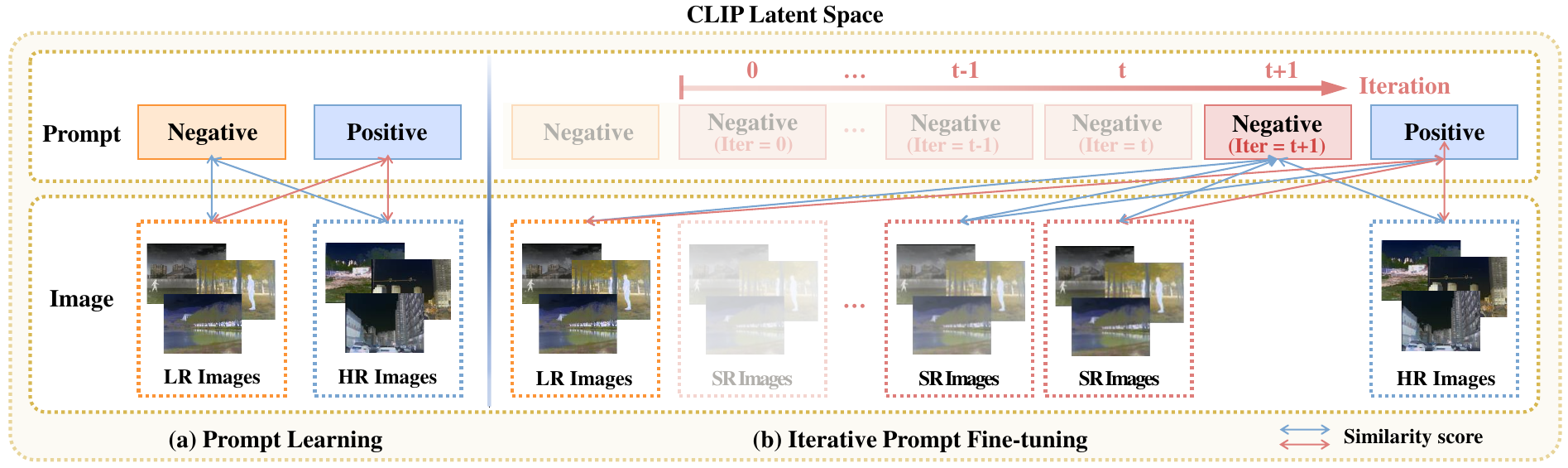}
    \caption{Overview of the prompt learning process. (a). The unlocked text encoder optimizes the learnable prompts to maximize the distance between negative and positive semantics in the latent space. (b). The degradation loss guides the SR process to align with positive prompts while distancing from the negative ones with the locked text encoder.}
    \label{pics:prompt learning process}
\end{figure*}

Once the local tokens are generated, they undergo a multi-head self-attention (MSA) layer followed by a multi-layer perceptron (MLP) to model global interactions as \(\mathbf{C}' = \mathbf{C} + \gamma_1 \cdot \text{MSA}(\text{LN}(\mathbf{C}))\) and \(\mathbf{C}'' = \mathbf{C}' + \gamma_2 \cdot \text{MLP}(\text{LN}(\mathbf{C}'))\), where $\gamma_1$ and $\gamma_2$ are learnable scaling factors and $\text{LN}$ represents layer normalization. This enables the model to capture long-range dependencies across windows. Next, the local tokens are concatenated with the window tokens to integrate the global context as \(\mathbf{X}_{w} = \text{Concat}(\mathbf{X}_{l}, \mathbf{C})\). The concatenated tokens then go through another attention layer to jointly model local and global information as \(\mathbf{X}_{w}' = \mathbf{X}_{w} + \gamma_3 \cdot \text{MSA}(\text{LN}(\mathbf{X}_{w}))\) and \(\mathbf{X}_{w}'' = \mathbf{X}_{w}' + \gamma_4 \cdot \text{MLP}(\text{LN}(\mathbf{X}_{w}'))\).

Subsequently, the local tokens and window tokens are separated again for subsequent attention layers as \(\mathbf{X}_{l}, \mathbf{C} = \text{Split}(\mathbf{X}_{w}'')\). At the end of the GLIA block, an upsampling operation is applied to the local tokens, and they are merged back into the feature map to ensure global information is propagated, formally \(\mathbf{X}_{\text{out}} = \text{Upsample}(\mathbf{C}, \text{size} = \text{H} \times \text{W}) + \mathbf{X}_{l}\). The proposed CRG framework enables the model to better perceive infrared characteristics, allowing for better extraction and restoration of high-frequency features in infrared.

\subsection{Loss functions}
\noindent \textbf{Prompt Learning Based Degradation Loss.}
To better learn the infrared-specific degradation pattern, we introduce a two-stage prompt learning strategy by leveraging the capabilities of the CLIP model. As shown in Figure~\ref{pics:cover} (e), this strategy augments the capacity to adaptively interpret and improve upon the quality of SR and details within reconstructed scenes.

The foundation of this prompt learning strategy is the prompt-based degradation loss, which utilizes the CLIP model's proficiency in semantic parsing to align the generated images with textual descriptors. We apply the three pairs of prompts to represent degradation specifically common in infrared images and formulate the degradation loss using CLIP to guide the super-resolution process toward generating images that semantically align with positive textual descriptors while distancing from the negative ones. 

As shown in Figure ~\ref{pics:prompt learning process}, our method initializes the prompt pairs from high-resolution (HR) and low-resolution (LR) image counterparts. The HR image undergoes encoding via the locked image encoder \(\mathbf{\Phi_{\text{image}}}\) of the CLIP model, producing its latent representation. Concurrently, the latent codes for the dichotomous prompts are derived through the unlocked text encoder \(\mathbf{\Phi_{\text{text}}}\). Leveraging the latent space similarity metric \(\text{SIM}(\mathbf{I, T}) = e^{\text{cos}(\mathbf{\Phi_{\text{image}}(I), \Phi_{\text{text}}(T))}}\), we apply binary cross-entropy loss to fine-tune the initial prompt pair, distinguishing HR from LR images:
\begin{equation}
\mathcal{L} =-(y * \log (\hat{y})+(1-y) * \log (1-\hat{y})),
\end{equation}
\begin{equation}
\hat{y} = \frac{\text{SIM}(\mathbf{\Phi}_{\text{image}}(\mathbf{I}), \mathbf{\Phi}_{\text{text}}(\mathbf{T}_{\text{pos}}))}{\sum_{i \in \{\text{neg}, \text{pos}\}} \text{SIM}(\mathbf{\Phi}_{\text{image}}(\mathbf{I}_i), \mathbf{\Phi}_{\text{text}}(\mathbf{T}_i))},
\end{equation}
\noindent where \(\mathbf{I}\) signifies the paired HR and LR images, and \(y\) is their corresponding label, designated 0 for LR and 1 for HR. \( \mathbf{T}_{\text{pos}} \) and \( \mathbf{T}_{\text{neg}} \) encapsulate the encoded features of the positive and negative prompts, respectively. 

After the initial stage of prompt optimization, we then lock the text encoder \(\mathbf{\Phi_{\text{text}}}\) and advance to refine our network with the degradation loss. The degradation loss \( \mathcal{L}_{\text{degrad}} \) for a batch of SR images \( \mathbf{I} \) is computed as:
\begin{equation}
    \mathcal{L}_{\text{degrad}} = \frac{1}{N} \sum_{i=1}^{N} \frac{\text{SIM}(\mathbf{\Phi}_{\text{image}}(\mathbf{I}_i), \mathbf{\Phi}_{\text{text}}(\mathbf{T}_{\text{neg}}))}{\text{SIM}(\mathbf{\Phi}_{\text{image}}(\mathbf{I}_i), \mathbf{\Phi}_{\text{text}}(\mathbf{T}_{\text{pos}}))},
\end{equation}
\noindent where \(N\) indicates the batch size. This loss function motivates the network to yield infrared degradation-aligned images with high-quality descriptors while diverging from the qualities of the low-quality ones, thus ensuring a visual alignment. The training alternates between refining the prompts and fine-tuning the enhancement network until the outputs achieve visual excellence. 

\noindent \textbf{Spectral Fidelity Loss.}
To better learn the infrared LR-HR image mapping, we propose a spectral fidelity loss that constrains the distribution of high- and low-frequency thermal radiation, ensuring the fidelity of infrared-specific features. Given the HR image \( \mathbf{I}_{\text{HR}} \) and the SR image \( \mathbf{I}_{\text{SR}} \), we begin by computing their 2D Discrete Fourier Transforms (DFT). This step transforms the spatial domain representation into the frequency domain, formally:

\begin{equation}
\begin{gathered}
    \hat{\mathbf{I}}_{\text{HR}} = \mathcal{F}(\mathbf{I}_{\text{HR}}),\\
    \mathcal{F}(u, v) = \sum_{x=0}^{H-1} \sum_{y=0}^{W-1} I(x, y) \cdot e^{-i \frac{2\pi}{H} ux} \cdot e^{-i \frac{2\pi}{W} vy}.
\end{gathered}
\end{equation}

\begin{figure*}[!hpt]
    \centering
    \includegraphics[width=1.0\linewidth]{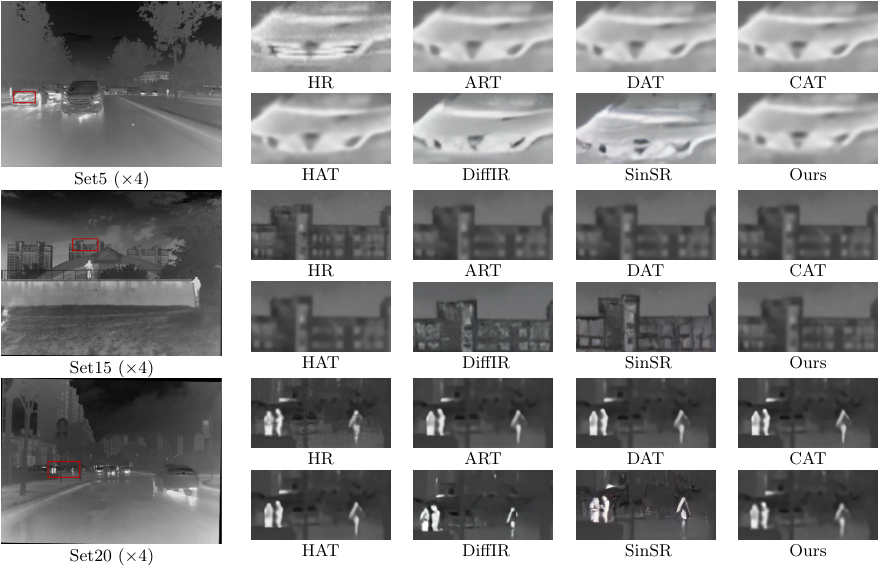}
    \caption{Visual comparison of infrared image SR ($\times$4) with SOTA methods.}
    \label{fig:sr_vs_2}
\end{figure*}

\noindent where \( \mathcal{F}(u, v) \) denotes the DFT of the image at frequency coordinates \( (u, v) \), and \( \hat{\mathbf{I}}_{\text{HR}} \) is the transformed HR images. The transformed SR image \( \hat{\mathbf{I}}_{\text{SR}} \) is computed the in the same way as \(\mathcal{F}(\mathbf{I}_{\text{SR}})\). Once in the frequency domain, we shift the zero-frequency component, corresponding to the image’s mean intensity, to the center of the spectrum for both the HR and SR images, resulting in \( \hat{\mathbf{I}}_{\text{HR}}^{\text{shift}} \) and \( \hat{\mathbf{I}}_{\text{SR}}^{\text{shift}} \), respectively. Next, the magnitude spectra of the Fourier-transformed images \( \mathbf{M}_{\text{HR}} \) and \( \mathbf{M}_{\text{SR}} \) are computed by applying a logarithmic compression. This ensures that both high-frequency and low-frequency components are effectively considered during the comparison. 

We then normalize the magnitude spectra to have zero mean and unit variance, denoted as \(\mathbf{M}_{\text{HR}}^{\text{norm}}\) and \(\mathbf{M}_{\text{SR}}^{\text{norm}}\), to ensure that the loss focuses on matching the frequency distribution patterns between the HR and SR images rather than on absolute intensity differences. Finally, the Spectral Fidelity Loss is computed as the mean squared error between the normalized magnitude spectra of the HR and SR images:
\begingroup
\small
\begin{equation}
\mathcal{L}_{\text{SF}}= \biggl( \overbrace{\mathcal{N}\bigl(\log(1 + |\hat{\mathbf{I}}_{\text{HR}}^{\text{shift}}|)\bigr)}^{\mathbf{M}_{\text{HR}}^{\text{norm}}} -  \overbrace{\mathcal{N}\bigl(\log(1 + |\hat{\mathbf{I}}_{\text{SR}}^{\text{shift}}|)\bigr)}^{\mathbf{M}_{\text{SR}}^{\text{norm}}} \biggr)^2,
\end{equation}
\endgroup
\noindent where \( \mathcal{N}(\cdot) \) represents the normalization operation. The Spectral Fidelity Loss plays a critical role in preserving the spectral distribution of the infrared image during the super-resolution process. By aligning the frequency components of the reconstructed image with the ground-truth HR image, the model ensures that crucial infrared-specific features, particularly those related to thermal radiation, are accurately restored. This preserves both the intricate high-frequency details and the smooth low-frequency information of the infrared image during super-resolution, resulting in an accurate reconstruction of the SR image. 

The total loss, \(\mathcal{L}_{\text{total}}\), integrates the degradation loss, spectral fidelity loss, with pixel and perceptual losses to collaboratively optimize the infrared images for thermal radiation distribution fidelity, visual fidelity, perceptual quality, and semantic congruence with high-quality image descriptions:
\begin{equation}
    \mathcal{L}_{\text{total}} = \mathcal{L}_{\text{SF}} + \mathcal{L}_{\text{degrad}} + \mathcal{L}_{\text{pixel}} + \mathcal{L}_{\text{perceptual}},
\end{equation}
\noindent where the pixel loss \(\mathcal{L}_{\text{pixel}}\), fundamentally an MSE calculation, assesses the pixel-level discrepancies between the super-resolved images and their high-resolution ground truth counterparts. Conversely, the perceptual loss \(\mathcal{L}_{\text{perceptual}}\) leverages a VGG network to extract and compare feature representations of the generated and ground truth images, focusing on minimizing differences in high-level feature representations. This comprehensive loss function ensures that our SR network reconstructs high-quality infrared images that not only closely resemble the ground truth but also align with the infrared degradation patterns, especially the thermal radiation distribution.

\section{Experiments}\label{sec4}
\subsection{Experimental Settings}
\textbf{Dataset and evaluation metrics.}
We assessed our model's performance against several benchmarks using publicly accessible infrared image datasets \(\text{M}^{3}\text{FD}\) ~\cite{liu2022target}, TNO~\cite{toet2017tno}, and RoadScene~\cite{xu2020u2fusion}. For training, we randomly selected 182 images from \(\text{M}^{3}\text{FD}\), while a set of 78 images from the same dataset was reserved for validation. We evaluate the SR performance using a mixture of five datasets: \(\text{M}^{3}\text{FD}\) \(\textbf{Set5}\), \(\textbf{Set15}\), and \(\textbf{Set20}\) along with RoadScene (60 images), and TNO (37 images). We conduct experiments with upscaling factors of \(\times 2\) and \(\times 4\), where LR images are derived from HR counterparts through bicubic degradation. To quantitatively assess the performance of our model, we utilize three widely used metrics, PSNR, MSE, and SSIM. The detailed information related to those metrics can be found here~\cite{wang2004image}. 

\noindent\textbf{Implementation Details.}
Our network was trained on a GeForce RTX 4090 GPU, utilizing the Adam optimizer for parameter updates. We set the initial learning rate to \(1e^{-4}\), employing an exponential decay strategy to refine the learning process over time. The training was executed with patch size 64\(\times\)64 and batch size 32. To enhance the robustness and generalizability of our model, we employed random rotations ($90^\circ$, $180^\circ$, $270^\circ$) and horizontal flips.

\begin{figure*}[!ht]
    \centering
    \includegraphics[width=1.0\linewidth]{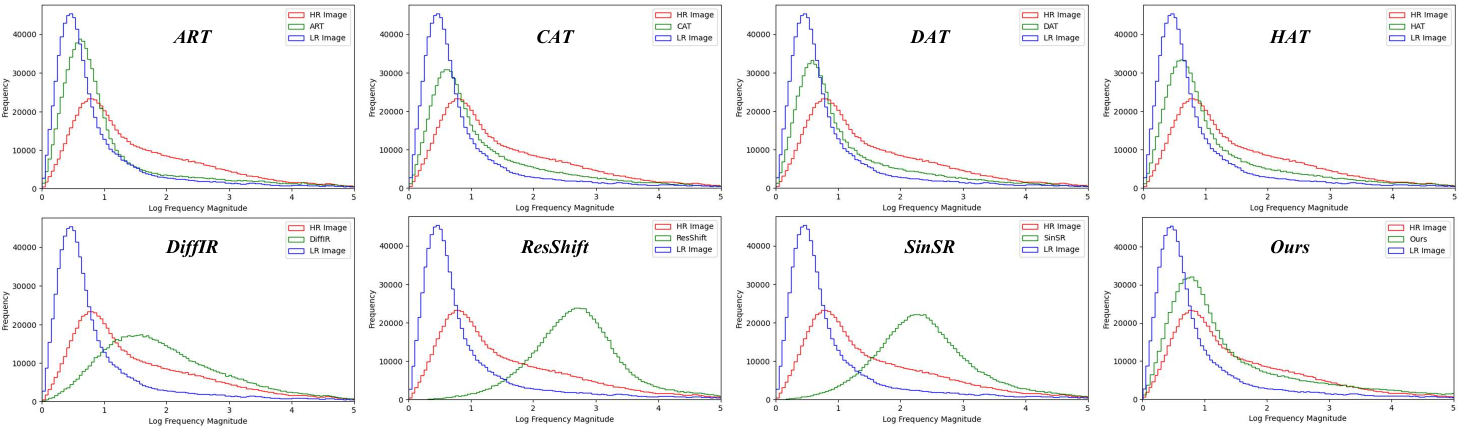}
    \caption{Visual comparison of thermal radiation distribution with SOTA methods.}
    \label{fig:distribution_compare}
\end{figure*}

\subsection{Experiments on Infrared SR}
We conduct a comprehensive comparison of our approach against nine state-of-the-art methods, including SwinIR ~\cite{liang2021swinir}, DAT ~\cite{chen2023dual}, CAT ~\cite{chen2022cross}, ART ~\cite{zhang2022accurate}, HAT ~\cite{chen2023activating}, EDT ~\cite{li2021efficient}, DiffIR~\cite{xia2023diffir}, SinSR~\cite{wang2024sinsr}, and ResShift~\cite{yue2024resshift} with their large, classic, and light versions if available.

\noindent\textbf{Quantitative Comparison.}
The quantitative super-resolution results of our method are presented in Table~\ref{tab:quantitative_results}, demonstrating its superior performance across various scales when compared with 18 state-of-the-art models on three representative sets from the \(\text{M}^{3}\text{FD}\). Notably, our approach consistently outperforms existing methods in terms of PSNR and SSIM, which are crucial for image quality and structural integrity. 

At a scale of \(\times 2\), our technique not only surpasses all classic and light models but also outperforms the majority of the larger versions of six cutting-edge methods, except for the MSE value on the \(\textbf{Set20}\). 
It trails only marginally behind EDT-B and EDT-B\textsuperscript{\dag}, marking a substantial advancement over its contemporaries. Specifically, compared with the EDT-B\textsuperscript{\dag}, the second best one of PSNR comparison on the \(\textbf{Set15}\) dataset (\(\times 2\)), our method achieves significant advancements, yielding 0.887 dB improvement. 

At a scale of \(\times 4\), our model secures the leading position in \(\textbf{Set15}\) and \(\textbf{Set20}\), underscoring our model's efficacy in reconstructing finer details and achieving higher fidelity in super-resolved images. All those results indicate that preserving the distribution fidelity of high- and low-frequency thermal radiation notably improves the infrared image SR quality.

\noindent\textbf{Qualitative Results.}
We further conduct a visual comparison on Figure~\ref{fig:sr_vs_2}. The \(\times 4\) SR results highlight the proficiency of our method in reducing artifacts while retaining spectral fidelity and finer high-frequency details. For instance, in the \(\textbf{Set5}\) results, competing methods frequently produce over-smoothed reconstructions that fail to preserve critical high-frequency details specific to infrared images. Also, a similar pattern is observed in the \(\textbf{Set20}\) results, where other models struggled with maintaining high-frequency details in complex infrared scenes. 

Moreover, diffusion-based methods, such as DiffIR and SinSR, tend to generate unexpected artifacts and alter the appearance of objects. As shown in \(\textbf{Set5}\) and \(\textbf{Set20}\), the car and pedestrians are barely recognizable. In contrast, our approach effectively preserves infrared details and textures, evidenced by sharper and more defined reconstructions.

\begin{table*}[!t]
\centering
  \scriptsize
  \renewcommand\arraystretch{1.1} 
	\setlength{\tabcolsep}{0.8mm}
  \caption{Quantitative comparison with the different SR methods on the \(\text{M}^{3}\text{FD}\) \ dataset at different scales. The best results are in \textbf{bold}, while the second-best results are \underline{underlined}.}
  \begin{tabular}{l|c|ccccccccc}
    \toprule
    \multicolumn{1}{l|}{\cellcolor{gray!10}}     & \cellcolor{gray!10}                        & \multicolumn{3}{c}{\cellcolor{gray!10}Set5} & \multicolumn{3}{c}{\cellcolor{gray!10}Set15} & \multicolumn{3}{c}{\cellcolor{gray!10}Set20}                                                                                                                                                                                                                                 \\
    \cellcolor{gray!10}\multirow{-2}{*}{Methods} & \cellcolor{gray!10}\multirow{-2}{*}{Scale} & \cellcolor{gray!10}PSNR$\uparrow$           & \cellcolor{gray!10}MSE $\downarrow$          & \cellcolor{gray!10}SSIM $\uparrow$           & \cellcolor{gray!10}PSNR $\uparrow$ & \cellcolor{gray!10}MSE $\downarrow$ & \cellcolor{gray!10}SSIM $\uparrow$ & \cellcolor{gray!10}PSNR $\uparrow$ & \cellcolor{gray!10}MSE $\downarrow$ & \cellcolor{gray!10}SSIM $\uparrow$ \\
    \midrule
    SwinIR-Lit                                 & x2                                         & 46.670                                      & 5.365                                        & 0.9896                                       & 47.120                             & 5.158                               & 0.9900                             & 47.680                             & 4.151                               & 0.9900                             \\
    SwinIR                                       & x2                                         & 46.880                                      & 5.121                                        & 0.9899                                       & 47.240                             & 5.001                               & 0.9902                             & 47.820                             & 4.023                               & 0.9901                             \\
    DAT-Lit                                    & x2                                         & 48.188                                      & 5.196                                        & 0.9920                                       & 48.555                             & 5.175                               & 0.9922                             & 49.093                             & 4.196                               & {0.9924}                             \\
    DAT-S                                        & x2                                         & 48.454                                      & 4.868                                        & 0.9919                                       & 48.825                             & 4.831                               & 0.9917                             & 49.366                             & 3.902                               & 0.9918                             \\
    DAT                                          & x2                                         & 48.434                                      & 4.818                                        & \underline{0.9922}                                     & 48.779                             & 4.800                               & \underline{0.9924}                           & 49.348                             & 3.848                               & \underline{0.9925}                           \\
    CAT-R-2                                      & x2                                         & 48.467                                      & 4.784                                        & 0.9911                                       & 48.817                             & 4.721                               & 0.9923                             & 49.422                             & 3.763                               & 0.9914                             \\
    CAT-R                                        & x2                                         & 48.490                                      & 4.710                                        & 0.9915                                       & 48.842                             & 4.652                               & 0.9923                             & 49.453                             & 3.707                               & 0.9918                             \\
    ART-S                                        & x2                                         & 48.368                                      & 5.858                                        & 0.9913                                       & 48.747                             & 5.214                               & 0.9923                             & 49.361                             & 5.705                               & 0.9916                             \\
    ART                                          & x2                                         & 48.470                                      & 5.096                                        & 0.9917                                       & 48.834                             & 4.479                               & \underline{0.9924}                             & 49.443                             & 4.061                               & \underline{0.9925}                             \\
    HAT-S                                        & x2                                         & 48.456                                      & 4.805                                        & 0.9919                                       & 48.841                             & 4.740                               & 0.9914                             & 49.416                             & 3.811                               & 0.9918                             \\
    HAT                                          & x2                                         & 48.430                                      & 4.878                                        & 0.9921                                       & 48.790                             & 4.805                               & 0.9923                             & 49.368                             & 3.867                               & {0.9924}                             \\
    HAT-L                                        & x2                                         & \underline{48.584}                                      & \underline{4.474}                                        & 0.9921                                       & 48.974                             & 4.418                               & \underline{0.9924}                             & 49.518                             & 3.561                               & \underline{0.9925}                             \\
    EDT-T                                        & x2                                         & 48.115                                      & 4.837                                        & 0.9917                                       & 48.601                             & 4.598                               & 0.9920                             & 49.104                             & 3.797                               & 0.9921                             \\
    EDT-B                                        & x2                                         & 48.188                                      & 4.497                                        & 0.9919                                       & 48.709                             & 4.392                               & 0.9921                             & 49.243                             & \underline{3.550}                               & 0.9922                             \\
    EDT-B\textsuperscript{\dag}                  & x2                                         & 48.570                                      & 4.527                                        & 0.9921                                       & \underline{49.008}                             & \underline{4.354}                               & \underline{0.9924}                             & \underline{49.558}                             & \textbf{3.503}                               & \underline{0.9925}                             \\
    DiffIR                                       & x2                                         & 37.192                                      & 69.398                                       & 0.9460                                       & 37.152                            & 63.971                              & 0.9468                             & 37.995                             & 47.342                              & 0.9562                             \\
    \midrule
    \textbf{Ours}                                & x2                                         & \textbf{48.637}& \textbf{4.470}& \textbf{0.9940}& \textbf{49.895}& \textbf{4.223}& \textbf{0.9956}& \textbf{50.144}& 3.598& \textbf{0.9951}\\
    \midrule
    SwinIR-Lit                                 & x4                                         & 38.720                                      & 35.478                                       & 0.9502                                       & 38.780                             & 38.190                              & 0.9473                             & 40.070                             & 24.694                              & 0.9620                             \\
    SwinIR                                       & x4                                         & 39.010                                      & 33.155                                       & 0.9524                                       & 40.120                             & 36.662                              & \underline{0.9598}                             & 41.680                             & 22.877                              & 0.9700                             \\
    DAT-Lit                                    & x4                                         & 40.345                                      & 32.555                                       & 0.9606                                       & 40.399                             & 35.299                              & 0.9574                             & 41.637                             & 23.078                              & 0.9701                             \\
    DAT-S                                        & x4                                         & 40.679                                      & \underline{30.101}                                       & 0.9622                                       & 40.702                             & 32.518                              & 0.9593                             & 41.919                             & 21.232                              & 0.9709                             \\
    DAT                                          & x4                                         & \underline{40.764}                                      & \textbf{29.454}                                       & \underline{0.9625}                                       & \underline{40.812}                             & 31.854                              & 0.9597                             & \underline{42.046}                             & 20.783                              & 0.9712                             \\
    CAT-R-2                                      & x4                                         & 40.580                                      & 30.675                                       & 0.9617                                       & 40.621                             & 32.879                              & 0.9588                             & 41.940                             & 21.139                              & 0.9708                             \\
    CAT-R                                        & x4                                         & 40.604                                      & 30.786                                       & 0.9619                                       & 40.612                             & 33.615                              & 0.9585                             & 41.879                             & 21.390                              & 0.9706                             \\
    ART-S                                        & x4                                         & 40.597                                      & 30.604                                       & 0.9615                                       & 40.626                             & \underline{30.250}                              & 0.9586                             & 41.917                             & 33.101                              & 0.9708                             \\
    ART                                          & x4                                         & 40.696                                      & 32.469                                       & 0.9619                                       & 40.754                             & 31.241                              & 0.9593                             & 42.045                             & \underline{20.702}                              & 0.9712                             \\
    HAT-S                                        & x4                                         & 40.677                                      & 30.536                                       & 0.9620                                       & 40.760                             & 32.389                              & 0.9594                             & 41.994                             & 21.191                              & 0.9710                             \\
    HAT                                          & x4                                         & 40.736                                      & 30.118                                       & 0.9622                                       & 40.740                             & 32.487                              & 0.9594                             & 41.996                             & 20.991                              & 0.9711                             \\
    HAT-L                                        & x4                                         & 40.754                                      & 30.467                                       & 0.9624                                       & 40.741                             & 31.985                              & 0.9595                             & 41.997                             & 20.753                              & \underline{0.9713}                             \\
    EDT-T                                        & x4                                         & 40.136                                      & 34.249                                       & 0.9532                                       & 40.383                             & 34.953                              & 0.9523                             & 41.259                             & 23.516                              & 0.9621                             \\
    EDT-B                                        & x4                                         & 40.520                                      & 31.501                                       & 0.9610                                       & 40.641                             & 33.549                              & 0.9584                             & 41.877                             & 21.817                              & 0.9706                             \\
    EDT-B\textsuperscript{\dag}                  & x4                                         & 40.618                                      & 30.606                                       & 0.9616                                       & 40.715                             & 32.497                              & 0.9590                             & 41.948                             & 21.041                              & 0.9708                             \\
    DiffIR                                       & x4                                         & 33.629                                      & 158.710                                      & 0.8992                                       & 33.981                            & 134.020                             & 0.9065                             & 34.672                             & 109.699                             & 0.9207                             \\
    SinSR                                        & x4                                         & 31.645                                      & 239.575                                      & 0.8481                                       & 31.988                             & 220.675                             & 0.8426                             & 33.438                             & 145.241                             & 0.8853                             \\
    ResShift                                     & x4                                         & 30.836                                      & 308.171                                      & 0.8179                                       & 31.253                             & 288.000                             & 0.8456                             & 31.375                             & 239.573                             & 0.8501                             \\
    \midrule
    \textbf{Ours}                                & x4                                         & \textbf{40.784}& 31.217& \textbf{0.9774}& \textbf{42.741}& \textbf{30.159}& \textbf{0.9781}& \textbf{42.734}& \textbf{20.248}& \textbf{0.9801}\\
    \bottomrule
  \end{tabular}
  \label{tab:quantitative_results}%
\end{table*}%

The experiment shown in Figure~\ref{fig:distribution_compare} further intuitively witnesses the spectral frequency distribution fidelity of infrared images reconstructed by our model. We visualize the log frequency magnitudes and their corresponding frequency by plotting the thermal radiation distribution histogram. The x-axis represents the spatial frequency components of the image, with higher values indicating higher spatial frequencies. The y-axis represents the occurrence of each frequency component, essentially the number of pixels that exhibit that spatial frequency in the image. 

Notably, our model closely aligns with the HR image, preserving the spectral frequency fidelity effectively. In contrast, other methods either significantly distort the frequency distribution, as seen with diffusion-based approaches, or shift the curve to the left, failing to retain high-frequency infrared-specific features. This distribution distortion explains why diffusion-based SR methods often lower the PSNR and MSE of the reconstructed images, despite their visually appealing sharpness. For instance, the SinSR tends to distort the thermal area of human to an arrow shape with their head and arms tightened together. The same pattern is also recognized in the DiffIR which twists the background buildings and creates artifacts around the ambiguous areas.

\subsection{Experiments on Infrared Object Detection}
We evaluate our model's ability to enhance infrared object detection performance using \(\text{M}^{3}\text{FD}\) detection dataset comprising six categories: Lamp, Car, Bus, Motor, Truck, and People. The assessment metric is mAP@.5:.95, representing the mean Average Precision across different IoU thresholds ranging from 0.5 to 0.95. To ensure a fair comparison, we directly input the reconstructed SR images of various SR models into the YOLOv5\footnote{https://github.com/ultralytics/yolov5} for retraining. The detector used in all experiments follows its original settings and the quantitative results are directly output by its test code for all experiments.

\begin{figure*}[!t]
    \centering
    \includegraphics[width=1.0\linewidth]{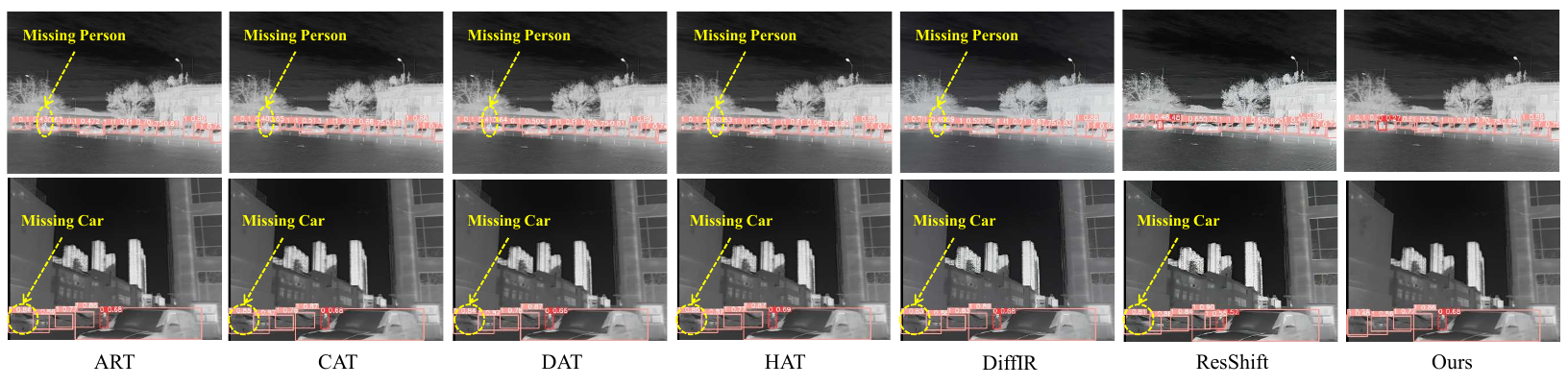}
    \caption{Visual comparison of object detection results with SOTA methods using YOLOv5.}
    \label{fig:yolov5_compare}
\end{figure*}

\begin{table*}[!t]
    \centering
    \scriptsize 
    \renewcommand\arraystretch{1.1}
    \setlength{\tabcolsep}{0.45mm}
    \caption{Quantitative comparison of object detection results using YOLOv5 at different scales. The best results are in \textbf{bold}, while the second-best results are \underline{underlined}.}
    \begin{tabular}{l|c|ccccccc|c|ccccccc}
        \toprule
                                    &                   & People            & Car            & Lamp              & Bus               & Motor             & Truck    & mAP         &         & People            & Car               & Lamp              & Bus               & Motor             & Truck        & mAP          \\
        \midrule
        {SwinIR-Lit}               & x2              & 0.391             & 0.537          & 0.185             & 0.611             & 0.190             & 0.399     & 0.385        & x4              & 0.378             & 0.518             & 0.168             & \textbf{0.596}    & 0.182             & 0.382    & 0.371         \\
        SwinIR                      & x2             & 0.392             & 0.537          & 0.182             & 0.611             & 0.191             & 0.402     & 0.386         & x4             & \textbf{0.383}    & 0.517             & 0.175             & 0.585             & 0.170             & 0.383    & 0.369          \\
        DAT-Lit                   & x2              & 0.391             & 0.536          & 0.184             & \underline{0.613} & 0.192         & 0.401       & 0.386      & x4            & 0.378             & 0.518             & 0.170             & \underline{0.593} & 0.171             & 0.390     & 0.370          \\
        DAT-S                       & x2              & 0.391             & 0.538          & 0.187 & \textbf{0.615}    & 0.191             & 0.397     & 0.386        & x4             & 0.380             & 0.520             & \underline{0.181} & 0.587             & 0.171             & \underline{0.394} & 0.372 \\
        DAT                         & x2  & \textbf{0.394}    & 0.536          & 0.186             & 0.612             & 0.195 & 0.404     & \underline{0.388}        & x4              & 0.380             & 0.520             & 0.177             & 0.585             & 0.179             & 0.391      & 0.372       \\
        CAT-R-2                     & x2              & 0.392             & 0.538          & 0.187 & \underline{0.613} & 0.189             & 0.401     & 0.387        & x4              & 0.381             & 0.517             & 0.179             & 0.583             & 0.176             & 0.387      & 0.370       \\
        CAT-R                       & x2              & 0.392             & 0.537          & 0.184             & \underline{0.613} & 0.191             & 0.401    & 0.386         & x4              & 0.381             & 0.519             & 0.178             & 0.581             & 0.174             & 0.389      & 0.370       \\
        ART-S                       & x2             & 0.392             & 0.538          & \underline{0.188}    & 0.608             & 0.195 & 0.399    & 0.387          & x4              & 0.381             & \underline{0.521} & 0.178             & 0.586             & 0.174             & 0.392      & 0.372       \\
        ART                         & x2             & 0.392             & \underline{0.539} & 0.187 & \underline{0.613} & 0.192             & 0.399    & 0.387          & x4  & 0.379             & 0.520             & 0.180             & 0.592             & \underline{0.183} & 0.390       & \underline{0.374}      \\
        HAT-S                       & x2              & \underline{0.393} & 0.538          & 0.184             & 0.610             & 0.191             & 0.399      &0.386        & x4             & 0.380             & 0.519             & 0.179             & 0.591             & 0.173             & 0.391      & 0.372        \\
        HAT                         & x2             & \underline{0.393} & 0.536          & 0.185             & \underline{0.613} & 0.192             & 0.400   & 0.386           & x4              & 0.380             & 0.519             & 0.178             & 0.584             & 0.180             & 0.387      & 0.371       \\
        HAT-L                       & x2              & 0.392             & 0.537          & 0.185             & \textbf{0.615}    & \underline{0.196}    & 0.399   & 0.387          & x4              & \textbf{0.383}    & 0.517             & \underline{0.181} & 0.585             & 0.177             & 0.393       & 0.373      \\
        EDT-T                       & x2              & 0.392             & 0.537          & 0.187 & 0.611             & 0.187             & \underline{0.406} & 0.387  & x4              & 0.379             & 0.519             & 0.172             & \underline{0.593} & 0.168             & 0.379       & 0.368      \\
        EDT-B                       & x2             & 0.392             & 0.536          & 0.187 & \underline{0.613} & 0.190             & 0.404     & 0.387         & x4              & \underline{0.382} & 0.519             & 0.176             & 0.584             & 0.178             & 0.376      & 0.369       \\
        EDT-B\textsuperscript{\dag} & x2              & 0.391             & 0.537          & 0.185             & 0.610             & \underline{0.196}    & 0.402      &0.387        & x4              & \textbf{0.383}    & 0.519             & \underline{0.181} & 0.576             & 0.176             & 0.388     & 0.370        \\
        DiffIR                      & x2              & 0.381             & 0.526          & 0.164             & 0.572             & 0.170             & 0.370     & 0.364        & x4              & 0.358             & 0.487             & 0.141             & 0.538             & 0.156             & 0.330      & 0.335       \\
        SinSR                       & -  & -                 & -                 & -              & -                 & -                 & -                 & -                 & x4              & 0.361             & 0.479             & 0.126             & 0.514             & 0.157             & 0.312       & 0.325      \\
        ResShift                    & -  & -                 & -                 & -              & -                 & -                 & -                 & -                 & x4             & 0.364             & 0.499             & 0.148             & 0.529             & 0.155             & 0.341       & 0.339       \\
        \midrule
        Ours                        & x2     & \underline{0.393}             & \textbf{0.561} & \textbf{0.191}             & 0.509             & \textbf{0.220}             & \textbf{0.576}  & \textbf{0.408}  & x4     & 0.378             & \textbf{0.546}    & \textbf{0.183}    & 0.533             & \textbf{0.206}    & \textbf{0.552}   & \textbf{0.400} \\
        \bottomrule
    \end{tabular}

    \label{tab:yolov5_results}
\end{table*}

\noindent\textbf{Quantitative Comparison.}  
As shown in Table~\ref{tab:yolov5_results}, our model achieves overall the best mAP across six recognition categories for both \(\times 2\) and \(\times 4\) scales, demonstrating a significant enhancement in machine perception for detection tasks. Notably, our model excels in detecting Car and Truck, with improvements of 0.17 and 0.158 in the Truck category for \(\times 2\) and \(\times 4\) scales, respectively. This indicates that preserving infrared-specific high-frequency information and maintaining the infrared spectral distribution is essential for improving object saliency and downstream machine perception, as reflected in both visually appealing reconstructed images and the increased detection accuracy.

\noindent\textbf{Qualitative Results.}  
The qualitative results shown in Figure~\ref{fig:yolov5_compare} visually emphasize the superiority of our model in object detection compared to competing methods. Notably, the other models frequently miss at least one label in complex infrared scenes. For example, only ResShift successfully detected the semi-occluded Person in the first row, while all other competing methods failed to do so. A similar pattern emerges in the second row, where all competing methods were unable to recognize the semi-occluded Car. In contrast, our model consistently identifies these labels in highly complex scenes, thanks to its enhanced ability to learn the LR-HR mapping and its efficient preservation of infrared-specific features, which are often overlooked by existing RGB image SR methods.

\begin{figure*}
    \centering
    \includegraphics[width=1.0\linewidth]{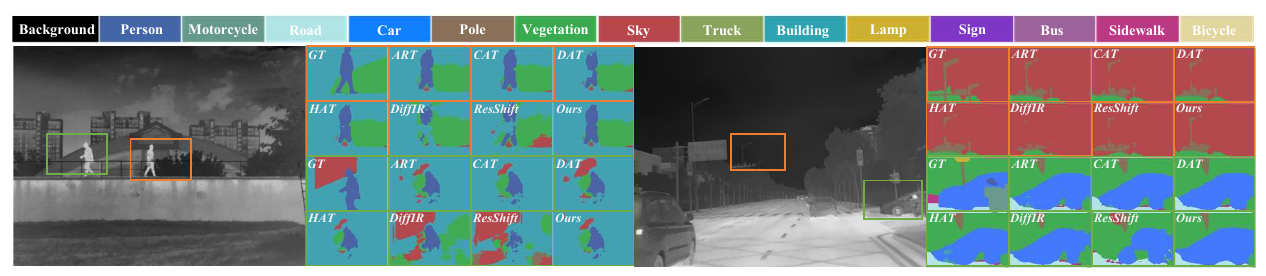}
    \caption{Visual comparison of semantic segmentation results with SOTA methods using SegFormer.}
    \label{fig:segformer_compare}
\end{figure*}

\begin{table*}[!pt]
    \centering
    \scriptsize
    \renewcommand\arraystretch{1.1}
    \setlength{\tabcolsep}{0.5mm}
    \caption{Quantitative comparison of semantic segmentation results using SegFormer at different scales. The best results are in \textbf{bold}, while the second-best results are \underline{underlined}.}
    \begin{tabular}{l|c|ccccccc|c|ccccccc}
        \toprule
                                    &    & Road              & Bud          & Tree              & Sky               & Hum             & Car               & mIoU              &    & Road              & Bud          & Tree              & Sky               & Hum             & Car               & mIoU              \\
        \midrule
        {SwinIR-Lit}                & x2 & 80.13             & 69.96             & 71.75             & 86.75             & 47.59             & 70.88             & 56.55             & x4 & 78.35             & 68.38             & 70.36             & 86.53             & 49.04             & 66.64             & 55.41             \\
        SwinIR                      & x2 & 80.24             & 69.92             & \textbf{71.83}    & 86.76             & 47.54             & 70.79             & 56.56             & x4 & 78.59             & 68.47             & 70.54             & 86.56             & 48.72             & 66.79             & 55.51             \\
        DAT-Lit                   & x2 & 80.18             & 69.87             & 71.67             & 86.72             & 47.59             & 70.75             & 56.52             & x4 & 78.66             & 68.58             & 70.54             & 86.66             & 49.49             & 66.62             & 55.60             \\
        DAT-S                       & x2 & 80.24             & 69.88             & 71.74             & 86.75             & 47.61             & 70.82             & 56.56             & x4 & 78.78             & 68.51             & 70.67             & 86.62             & 49.00             & 66.96             & 55.61             \\
        DAT                         & x2 & 80.24             & 69.92             & \underline{71.79} & \underline{86.78} & 47.41             & 70.82             & 56.55             & x4 & \underline{78.92} & \underline{68.78}    & \textbf{70.98}    & 86.67             & 48.87             & \underline{67.22}    & 55.73             \\
        CAT-R-2                     & x2 & 80.19             & 69.88             & 71.75             & 86.75             & 47.39             & 70.68             & 56.51             & x4 & 78.64             & 68.51             & 70.60             & 86.56             & 48.98             & 66.93             & 55.57             \\
        CAT-R                       & x2 & 80.23             & 69.87             & 71.74             & \underline{86.78} & 47.42             & 70.65             & 56.52             & x4 & 78.61             & 68.63             & 70.81             & 86.60             & 48.78             & 67.00             & 55.60             \\
        ART-S                       & x2 & 80.24             & 69.92             & 71.77             & \textbf{86.79}    & 47.52             & 70.79             & 56.56             & x4 & 78.65             & 68.38             & 70.51             & \underline{86.68} & 49.07             & 66.41             & 55.49             \\
        ART                         & x2 & \underline{80.35} & \underline{69.93} & 71.73             & \textbf{86.79}    & 47.53             & 70.94             & \underline{56.58} & x4 & 78.73             & 68.43             & 70.44             & 86.55             & 49.05             & 66.63             & 55.52             \\
        HAT-S                       & x2 & 80.23             & 69.84             & 71.74             & 86.76             & 47.50             & 70.72             & 56.52             & x4 & 78.78             & 68.59             & 70.76             & 86.67             & 48.59             & 66.63             & 55.55             \\
        HAT                         & x2 & 80.17             & 69.88             & 71.75             & 86.67             & 47.67             & 70.83             & 56.54             & x4 & 78.90             & 68.55             & 70.81             & 86.63             & 48.88             & 66.85             & 55.64             \\
        HAT-L                       & x2 & 80.19             & 69.87             & 71.70              & 86.62             & \underline{47.68} & 70.88             & 56.55             & x4 & \textbf{78.96}    & 68.54             & \underline{70.82} & 86.60             & 48.61             & 66.98             & 55.63             \\
        EDT-T                       & x2 & 79.84             & 69.68             & 71.49             & 86.28             & 47.59             & \textbf{71.01}    & 56.39             & x4 & 78.57             & 68.69             & 70.65             & \textbf{86.72}    & \underline{49.77} & 66.78             & 55.67             \\
        EDT-B                       & x2 & 79.86             & 69.63             & 71.47             & 86.25             & 47.61             & \textbf{71.01}    & 55.49             & x4 & 78.63             & 68.42             & 70.49             & 86.59             & 48.77             & 66.83             & \underline{56.39} \\
        EDT-B\textsuperscript{\dag} & x2 & 80.17             & 69.88             & 71.72             & 86.72             & 47.55             & 70.77             & 56.53             & x4 & 78.78             & 68.51             & 70.51             & 86.61             & 49.05             & 67.00             & 55.60             \\
        DiffIR                      & x2 & 76.53             & 66.62             & 67.44             & 81.94             & 46.54             & 68.28             & 49.90             & x4 & 73.16             & 61.76             & 58.79             & 80.46             & 46.18             & 59.88             & 53.62             \\
        SinSR                       & x2 & -                 & -                 & -                 & -                 & -                 & -                 & -                 & x4 & 73.78             & 63.61             & 64.61             & 80.86             & 44.51             & 58.42             & 50.75             \\
        ResShift                    & x2 & -                 & -                 & -                 & -                 & -                 & -                 & -                 & x4 & 74.46             & 63.06             & 62.12             & 79.92             & 48.65             & 62.98             & 51.61             \\
        \midrule
        Ours                        & x2 & \textbf{80.37}    & \textbf{69.98}    & 71.70             & \underline{86.78} & \textbf{47.74}    & \underline{70.98} & \textbf{57.24}    & x4 & \textbf{78.96}    & \textbf{68.79} & \underline{70.82}             & 86.59             & \textbf{50.18}    & \textbf{67.41} & \textbf{56.89}    \\
        \bottomrule
    \end{tabular}
    \label{tab:segformer_results}
\end{table*}

\subsection{Experiments on Infrared Semantic Segmentation}
We evaluate the infrared semantic segmentation performance on the FMB benchmark~\cite{liu2023multi} that contains 15 categories: Background, Person, Motorcycle, Road, Car, Pole, Vegetation, Sky, Truck, Building, Lamp, Sign, Bus, Sidewalk, and Bicycle. We select 5 out of 15 representative categories and show the IoU value of the selected major categories and the mIoU of all categories. For a fair comparison, we retrained the SegFormer~\cite{xie2021segformer} on the FMB dataset and then directly used the reconstructed SR image of various methods as inputs to inference. 

\noindent\textbf{Quantitative Comparison.} 
Table~\ref{tab:segformer_results} presents the quantitative comparison of infrared semantic segmentation results. Overall, our method outperforms the comparative methods on Road and Human categories and achieves the highest mIoU at both \(\times 2\) and \(\times 4\) scales. Notably, at the \(\times 2\) scale, our approach achieves the highest performance in four out of six categories and is ranked the second best in the other two categories, with the value of the Tree category trailing slightly behind the SwinIR. Similarly, at the \(\times 4\) scale, our method maintains the leading position in most categories, either the best or the second best, falling just 0.85 behind the HAT-L and 0.13 behind the top EDT-T. This demonstrates the superior capability of our model in maintaining high-frequency infrared-specific features, resulting in more distinguishable scenes.

\begin{figure*}[!htp]
    \centering
    \includegraphics[width=1.0\linewidth]{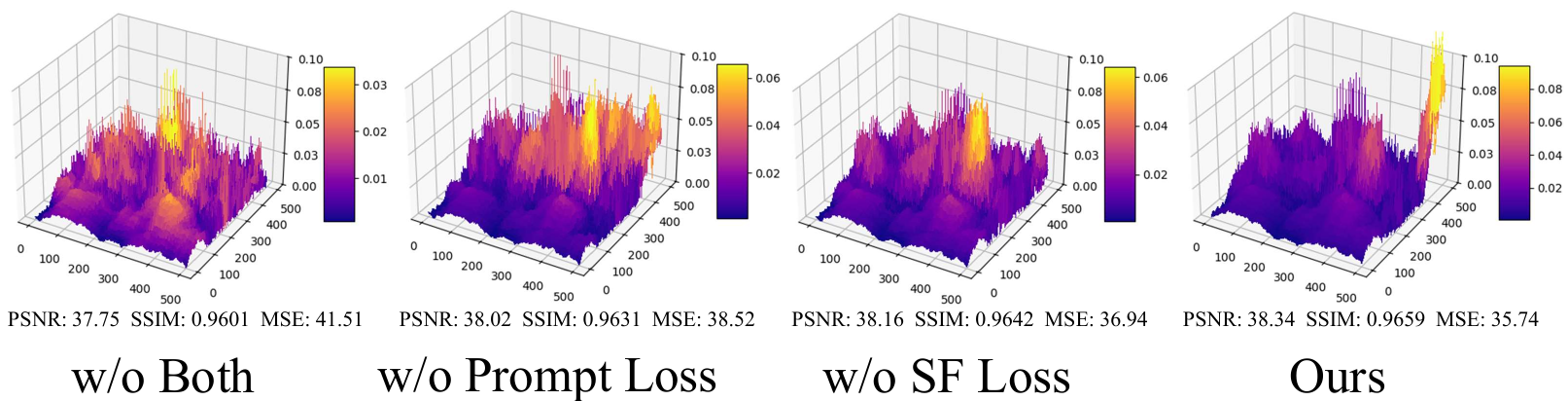}
    \caption{Visual ablation results of losses, showing the difference in pixel intensity with the HR images.}
    \label{fig:abla_prompt_sf_loss}
\end{figure*}

\noindent\textbf{Qualitative Results.}  
The qualitative results shown in Figure~\ref{fig:segformer_compare} demonstrate that existing RGB SR methods often fail to detect dim infrared targets or accurately identify distant pedestrians. For instance, all other methods failed to fully recognize the Pole on the right-hand side, with only ART coming close but still producing incorrect recognition. Similarly, on the left-hand side, all comparative methods struggled to identify the entire Person, either leaving it incomplete or misidentifying it. In contrast, our method successfully recognizes the full region with minimal errors, thanks to its ability to learn the infrared-specific LR-HR mapping effectively.

\begin{figure}[!t]
    \centering
    \includegraphics[width=1.0\linewidth]{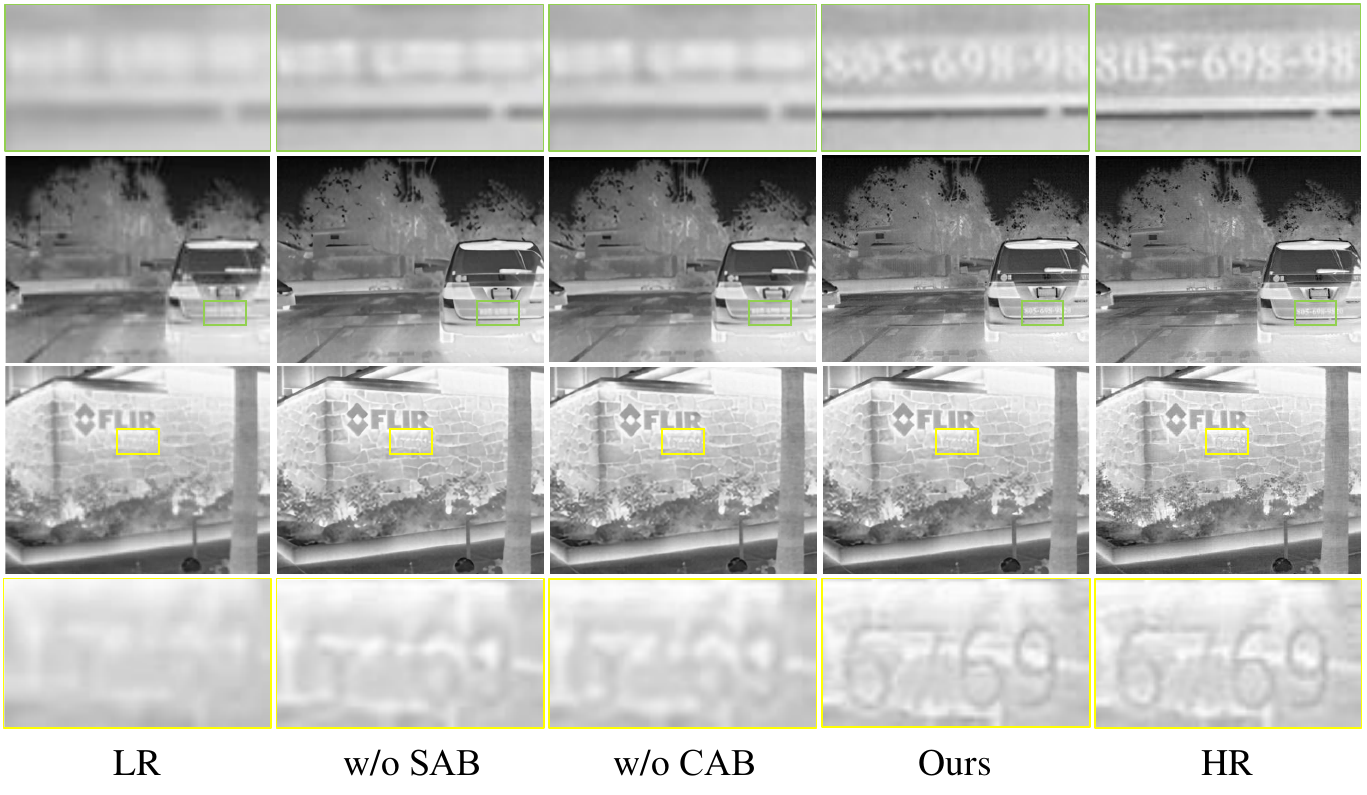}
     \caption{Visual ablation results ($\times$4) of different attention choices.}
    \label{abla:attention}
\end{figure}

\subsection{Ablation Studies}
\subsubsection{Experiments on Attention Blocks}
Figure~\ref{abla:attention} illustrates the individual and combined effects of the Spatial Attention Block (SAB) and Channel Attention Block (CAB) within our framework. The ablation results indicate that the combined use of SAB and CAB significantly improves the reconstruction of spatial features in infrared super-resolution tasks. Models without SAB experience a loss in preserving structural details, while the lack of CAB reduces the ability to capture textural nuances. For instance, the edges of the number shown in the figure are not well-preserved without either SAB or CAB. In contrast, our proposed model, which integrates both SAB and CAB, achieves results that closely approximate the high-resolution (HR) reference, effectively capturing fine details and maintaining the fidelity of the HR image. 

\begin{figure}
    \centering
    \includegraphics[width=1.0\linewidth]{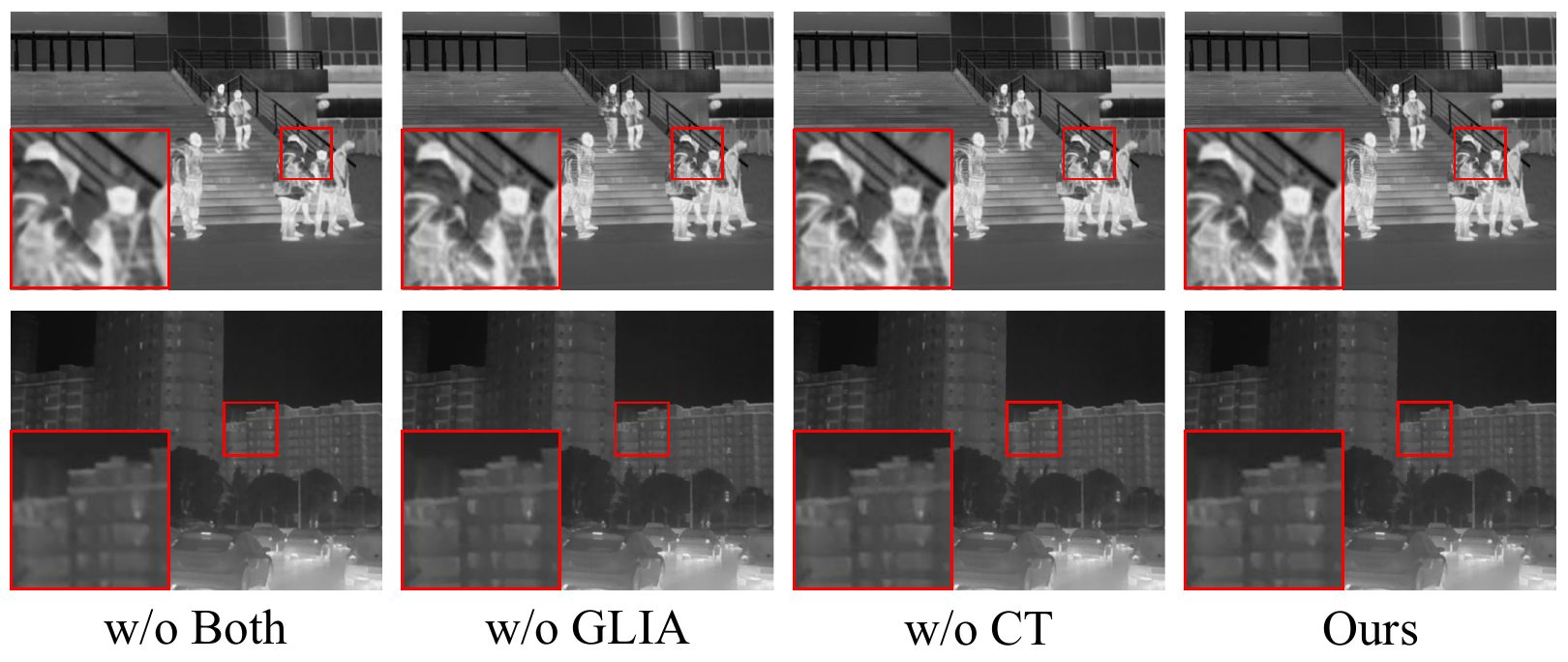}
    \caption{Visual ablation results of Contourlet Refinement Gate components. }
    \label{fig:abla_new_attention}
\end{figure}

\subsubsection{Study of Losses}
We draw Figure~\ref{fig:abla_prompt_sf_loss} to evaluate the effects of the degradation loss \(\mathcal{L}_{\text{degrad}}\) and spectral fidelity loss \(\mathcal{L}_{\text{SF}}\) on the infrared LR-HR mapping process. The figure illustrates the absolute pixel intensity differences between the HR image and SR images generated with both losses, one of the losses, and without any loss. Warmer colors indicate larger pixel intensity differences from the HR image. It is evident that omitting either loss results in a decline in the model's ability to maintain the overall pixel intensity distribution, deteriorating the PSNR, SSIM, and MSE values. In contrast, the model that combines both losses produces a mostly purple graph with the best SR results, indicating a close match between the SR and HR images in terms of pixel intensity. This demonstrates that both the degradation loss \(\mathcal{L}_{\text{degrad}}\) and spectral fidelity loss \(\mathcal{L}_{\text{SF}}\) are crucial for accurately learning the infrared LR-HR mapping. Since the degradation loss \(\mathcal{L}_{\text{degrad}}\) guides the network to learn infrared HR characteristics from LR degradation patterns, while the spectral fidelity loss \(\mathcal{L}_{\text{SF}}\) ensures the reconstructed infrared images close align with the HR images to maintain the infrared spectral distribution fidelity.

\subsubsection{Ablation of Contourlet Refinement Gate}
Figure~\ref{fig:abla_new_attention} presents a visual ablation study to assess the effectiveness of the proposed Contourlet Refinement Gate architecture. The Contourlet Transform (CT) block and the Global-Local Interactive Attention (GLIA) block are crucial components in enhancing the extraction of infrared-specific features. As shown in the first row of the figure, excluding both the CT and GLIA blocks results in SR images that exhibit noticeable blur and noise around thermal objects. The removal of either GLIA or CT individually leads to a degradation in fine background details, as illustrated in the second row, where the structure of the building appears slightly distorted. These results demonstrate that incorporating both the CT and GLIA blocks enables the model to reconstruct appealing infrared images by effectively capturing modal-specific features.

\begin{table}[!pt]
  \centering
  \scriptsize
  \setlength{\tabcolsep}{1.4mm}
  \caption{Ablation study of Contourlet Transform structure.}
  \begin{tabular}{cccc|ccc}
    \toprule
    \rowcolor{gray!10}\multicolumn{1}{l}{DFB} & \multicolumn{1}{l}{LP} & \multicolumn{1}{l}{PSNR} & \multicolumn{1}{l}{SSIM} & \multicolumn{1}{l}{L} & \multicolumn{1}{l}{PSNR} & \multicolumn{1}{l}{SSIM} \\
    \midrule
    &  &    40.031 & 0.9317 & 1 & 40.742 & \underline{0.9652} \\
    \checkmark  &     & 40.469 & 0.9588 & 2 & 40.755 & 0.9626 \\
          & \checkmark    & \underline{40.521} & \underline{0.9602} & 3 & \underline{40.770} & 0.9647 \\
    \checkmark     & \checkmark    & \textbf{40.787} & \textbf{0.9774} & 4 & \textbf{40.787} & \textbf{0.9774} \\
    \bottomrule
  \end{tabular}
  \label{aba_contourlet_combined}
\end{table}

\subsubsection{Examine Contourlet Transform Structure}
\begin{figure}
    \centering
    \includegraphics[width=0.47\textwidth]{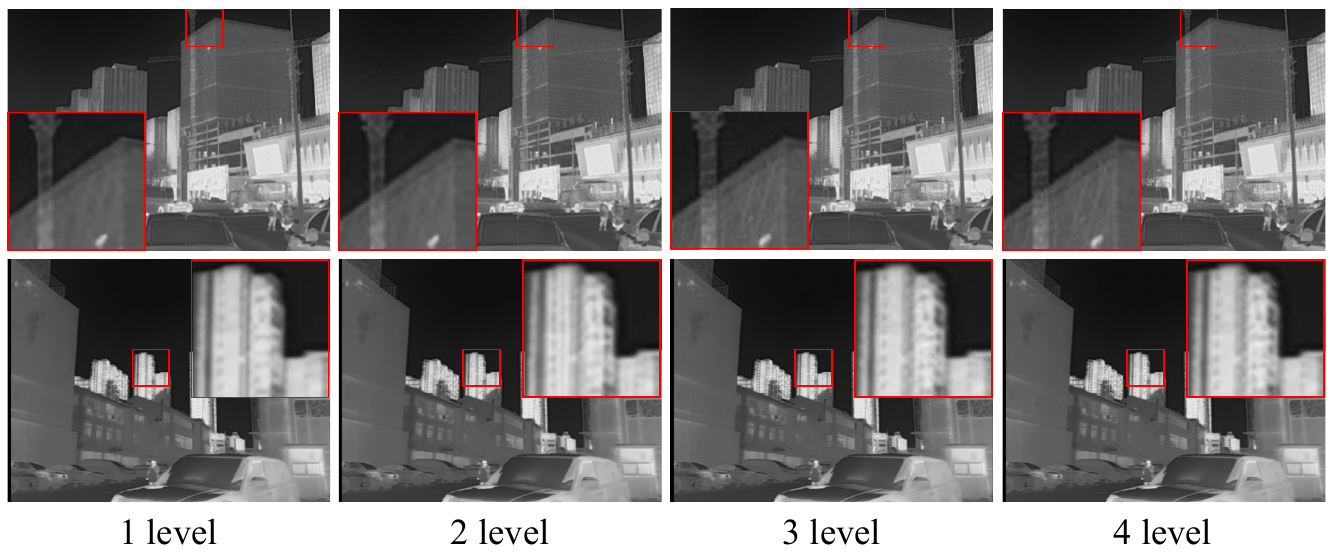}
    \captionof{figure}{Visual comparison of different contourlet decomposition levels.}
    \label{comparison other datasets}
\end{figure}

The results illustrated in Table~\ref{aba_contourlet_combined} on the left-hand side provide insights into the roles of the Laplacian Pyramid (LP) and Directional Filter Bank (DFB) within the Contourlet Transform (CT) block. Four configurations were tested: the absence of both DFB and LP, the combined use of DFB and LP, the exclusive use of DFB, and the sole use of LP. The results demonstrate the effectiveness of both DFB and LP, as their combined usage yields the highest PSNR, underscoring their complementary contributions in capturing multi-directional and multi-scale high-pass details in infrared images. The omission of either component leads to a deterioration in the restoration and enhancement of infrared-specific features from the multi-scale infrared spectral decomposition, ultimately compromising the quality of infrared image reconstruction. To further assess the impact of different levels in the Contourlet decomposition process on image super-resolution, Table~\ref{aba_contourlet_combined} right-hand side presents a detailed quantitative comparison. By increasing the number of Contourlet levels, more layers are introduced in the Laplacian Pyramid, along with corresponding directional filter banks at each layer. This multi-level approach allows the model to capture high-frequency details across a wider range of scales and orientations, thereby improving its ability to interpret the intricate characteristics of infrared spectra. The incremental rise in PSNR and SSIM with each additional level demonstrates the model's improved capability in resolving fine textures in infrared images. The visual ablation results in Figure~\ref{comparison other datasets} further confirm the effectiveness of increasing the decomposition levels.

\begin{table}[!pt]
    \centering
    \scriptsize
    \setlength{\tabcolsep}{1.8mm}
    \caption{Ablation study of prompt choice.}
    \begin{tabular}{ccccc}
    \toprule
    \rowcolor{gray!10}Baseline & Positive & Negative & PSNR  & SSIM \\
    \midrule
    \checkmark     &       &       & 40.431 & 0.9613 \\
    \checkmark     & \checkmark     &       & \underline{40.742} & \underline{0.9653} \\
    \checkmark     & \checkmark     & \checkmark     & \textbf{40.787} & \textbf{0.9774} \\
    \bottomrule
    \end{tabular}%
  \label{aba prompt choice}%
\end{table}

\begin{table*}[t]
    \begin{minipage}[c]{0.49\linewidth}
        \centering
        \includegraphics[width=1.0\textwidth]{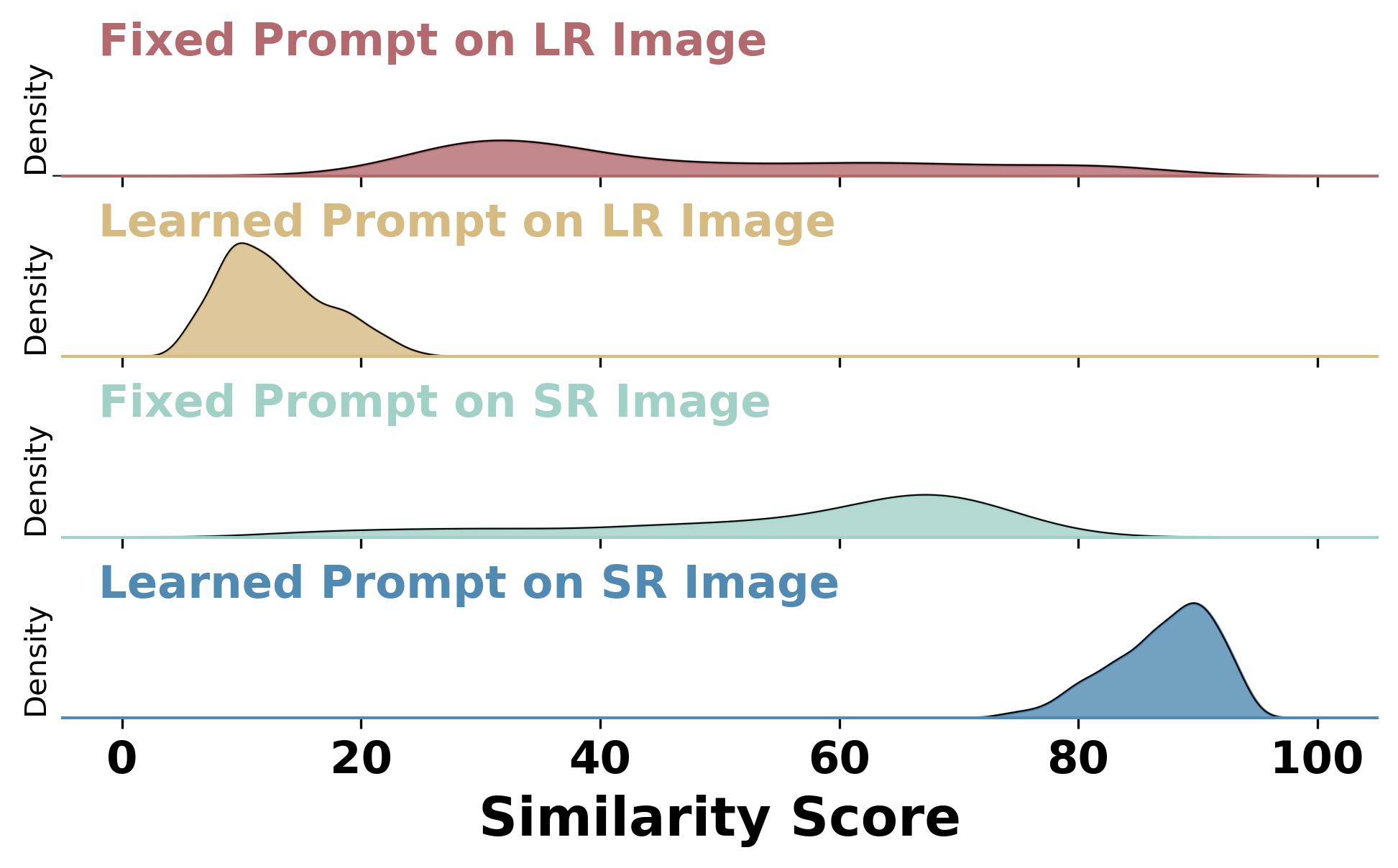}
       \captionof{figure}{Similarity comparison of ablation study on the prompt learning strategy.}
      \label{ablation_prompt_strategy_img}
    \end{minipage}\hfill
    \begin{minipage}[c]{0.48\linewidth}
        \centering
        \setlength{\tabcolsep}{1.0mm}
        \scriptsize
  \caption{Ablation study of different prompt strategies.}
    \begin{tabular}{ccc}
    \toprule
    \rowcolor{gray!10}Methods & PSNR  & SSIM \\
    \midrule
    Fixed prompts(Positive\&Negative) & 40.662 & 0.9614 \\
    Learned prompts(Negative) & \underline{40.725} & \underline{0.9628} \\
    Learned prompts(Positive) & 40.719 & 0.9622 \\
    Ours   & \textbf{40.787} & \textbf{0.9774} \\
    \bottomrule
    \end{tabular}%
  \label{aba prompt strategy}%
    \end{minipage}
\end{table*}

\subsubsection{Analyzing the Prompt Choice}
Table~\ref{aba prompt choice} highlights the influence of positive and negative prompts on the super-resolution process. Starting with the baseline model, which operates without prompts, the subsequent introduction of positive prompts led to a significant enhancement in both PSNR and SSIM, demonstrating the value of textual guidance that aligns with infrared-specific characteristics. When negative prompts were included alongside the positive ones, the model was further refined, moving closer to the desired outcomes and away from the negative influences. This refinement is reflected in the highest PSNR and SSIM scores, showcasing how prompts can effectively direct the model toward achieving a more accurate reconstruction of high-quality infrared images.

\subsubsection{Impact of Learning Strategy}
Figure~\ref{ablation_prompt_strategy_img} illustrates the impacts of the prompt learning strategy by showing the kernel density estimates of similarity scores between prompts and images on the \(\text{M}^{3}\text{FD}\) test dataset. The results reveal that learned prompts produce a sharper peak in similarity scores with super-resolved (SR) images compared to fixed prompts, indicating a stronger alignment with the SR outputs. In contrast, the broader distribution of fixed prompts suggests they capture the SR image features less accurately. Our two-stage prompt learning fine-tunes the initial prompt pair to distinguish between HR and LR images. Hence the degradation loss \(\mathcal{L}_{\text{degrad}}\) accurately regulates the reconstruction results. The noticeable shift towards higher similarity scores emphasizes the success of our two-stage learning approach in refining prompts to better represent high-resolution infrared image characteristics. As shown in Table~\ref{aba prompt strategy}, the learned prompts also maximize the separation between positive and negative samples in the CLIP latent space, ensuring more effective guidance of the model toward positive attributes while steering clear of negative ones.

\section{Conclusion}\label{sec5}
In this paper, we are the first to emphasize the infrared spectral distribution fidelity and propose an efficient paradigm for infrared image SR through improvements in infrared modal-specific feature extraction and infrared LR-HR mapping. Current SR techniques are predominantly optimized for visible light imaging, often overlooking the unique properties of infrared light. This results in distortions in the infrared spectrum distribution, ultimately compromising machine perception in downstream tasks. Our approach resolves these challenges by analyzing the distinctive spectral signatures of infrared images and efficiently restoring high-frequency details, thereby enhancing the learning of infrared LR-HR mappings and enabling more effective extraction of infrared-specific features. Specifically, we introduce a spectral fidelity loss that regularizes the thermal spectrum distribution during reconstruction, preserving both high- and low-frequency components to better capture the infrared LR-HR mapping. The proposed Contourlet Refinement Gate framework further enhances the extraction of modal-specific features through multi-scale multi-directional infrared spectral decomposition and global-local interactive attention. Our two-stage prompt learning strategy incrementally optimizes the model, aligning it more closely with the characteristics of infrared images through a learnable prompts strategy. Extensive experiments demonstrate that our method not only reconstructs visually appealing images but also notably enhances the machine perception at downstream tasks.

\section{Data Availability Statement}\label{sec6}
All code and data supporting the findings and experiments of this paper are publicly available in the GitHub repository at \href{https://github.com/hey-it-s-me/CoRPLE}{\texttt{https://github.com/hey-it-s-me/CoRPLE}}.
\balance

\end{document}